\documentclass[10pt,twocolumn,letterpaper]{article}

\usepackage[pagenumbers]{cvpr}      

\usepackage{microtype}

\renewcommand{\paragraph}[1]{\vspace{.5em}\noindent\textbf{#1.}}

\usepackage[T1]{fontenc}
\usepackage{textcomp,newunicodechar}

\usepackage{array}
\usepackage{graphicx} \usepackage{stfloats}
\newcolumntype{L}[1]{>{\raggedright\arraybackslash}p{#1}}

\definecolor{cvprblue}{rgb}{0.21,0.49,0.74}
\usepackage[pagebackref,breaklinks,colorlinks,allcolors=cvprblue]{hyperref}

\def\confName{CVPR}
\def\confYear{2027}

\def\mymethod{GeoFocus}
\usepackage{amsmath}
\usepackage{amssymb}
\usepackage[ruled,vlined,linesnumbered]{algorithm2e}
\usepackage{booktabs}   \usepackage{multirow}   \usepackage{pifont}     \usepackage{makecell}
\usepackage{xcolor}

\title{GeoFocus: Blending Efficient Global-to-Local Perception for Multimodal Geometry Problem-Solving}

\author{
Linger Deng\textsuperscript{1}\quad
Yuliang Liu\textsuperscript{1}\thanks{Corresponding author}\quad
Wenwen Yu\textsuperscript{1}\quad
Zujia Zhang\textsuperscript{1}\\
Pei Fu\textsuperscript{2}\quad
Zhenbo Luo\textsuperscript{2}\quad
Hao Lu\textsuperscript{1}\quad
Xiang Bai\textsuperscript{1}\\[0.5em]
{\small \textsuperscript{1}Huazhong University of Science and Technology}\\
{\small \textsuperscript{2}Xiaoai Plus, Xiaomi Inc}
}

\begin{document}
\maketitle
\begin{abstract}
    Geometry problem-solving (GPS) remains challenging for large multimodal models, partly due to limitations in geometric perception. Existing perception-oriented approaches face two key limitations: code-based diagram reconstruction relies on verbose training targets, while perception question-answering supervision may overlook local geometric relations critical to reasoning. To address these limitations, we propose \textbf{GeoFocus}, a global-to-local perception training framework with two complementary components. \textbf{VertexLang} provides compact supervision of global diagram topology through vertex coordinates and connectivity, reducing annotation length by \textbf{88\%} and training time by \textbf{20\%} compared with code-based reconstruction. The \textbf{Critical Local Perceptor} generates targeted question-answer supervision across \textbf{13 categories} of reasoning-relevant local geometric relations. On three GPS benchmarks, GeoFocus improves Qwen3-VL-2B/4B over matched GRPO-only baselines by \textbf{5.2/4.9 percentage points} on average, without external tools or additional inference steps. A human evaluation of 200 problems further shows that GeoFocus reduces cases with geometric perception errors from \textbf{118 to 80}. 
    These results highlight perception-focused training as an effective complement to reasoning optimization for multimodal geometric problem-solving.
\end{abstract}

\section{Introduction}

Although large multimodal models (LMMs) perform well on general visual tasks~\cite{liu2023visual,shen2021much}, they still struggle with geometry problem-solving (GPS), which requires accurately perceiving diagrams and reasoning about underlying relations. Recent methods enhance geometric reasoning through external tools~\cite{hu2026draw2think, yang2025bridging, zhang2026geomathcode, wang2026binsgps, li2026verifiable} or test-time scaling~\cite{kang2026reactivating, dong2026beyond}. However, accurate geometric perception remains challenging, as LMMs receive limited explicit supervision on the geometric structures and relations essential for GPS~\cite{zhang2024mathverse}.

\begin{figure}[t!]
    \centering
    \includegraphics[width=1\linewidth]{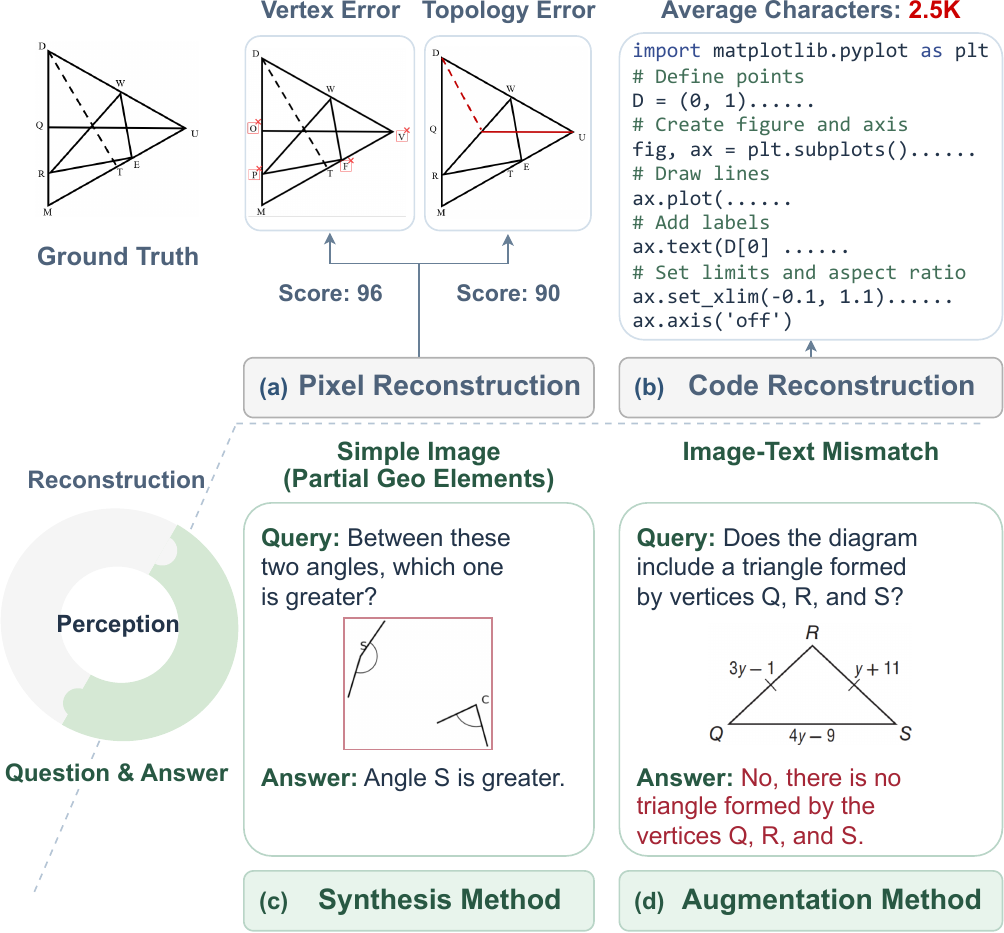}
    \caption{\textbf{Two main paradigms for enhancing geometry perception of LMMs}. One is image reconstruction (a,b); the other is perception Q\&A generation (c,d).}
    \label{fig:intro}
\end{figure}

A natural approach to improving geometric perception is to reconstruct diagrams, making global geometric structure an explicit learning target~\cite{cheng2025geouni,wang2025mathcoder}. Pixel-based reconstruction captures visual details but struggles to preserve vertex annotations and topological relations (Fig.~\ref{fig:intro}(a)). To better capture these relations, code-based methods represent diagrams through lengthy rendering programs, introducing substantial training overhead (Fig.~\ref{fig:intro}(b)). Beyond global reconstruction, another line of research focuses on learning geometric properties through perception question-answering (Q\&A). Image synthesis methods~\cite{zhang2024mavis,huang2025vision} provide accurate supervision but typically involve simple configurations, limiting coverage of complex relations such as midlines and angle bisectors (Fig.~\ref{fig:intro}(c)). To extend supervision to more complex diagrams, LMM-based augmentation methods~\cite{gao2025g,luo2025ursa} leverage existing geometry images, but may overlook critical local relations or introduce hallucinated descriptions (Fig.~\ref{fig:intro}(d)).

\begin{figure}[t!]
    \centering
    \includegraphics[width=1\linewidth]{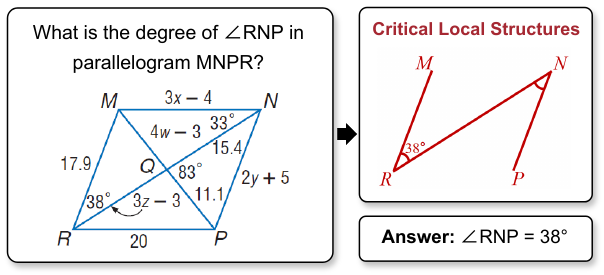}
    \caption{\textbf{GPS perception bottleneck}. Dense geometric annotations challenge global structure understanding and the identification of critical local relations essential for GPS.}
    \label{fig:intro_local}
\end{figure}

As illustrated in Fig.~\ref{fig:intro_local}, densely annotated diagrams challenge global geometric understanding and obscure local relations critical to reasoning.
However, existing methods often rely on verbose global reconstruction or provide insufficient supervision for reasoning-relevant local structures.
\textit{How can we establish accurate global geometric understanding and further identify the local relations critical to reasoning?}

To address these limitations, we treat geometric perception as two complementary learning targets: \textbf{global topology}, which establishes how geometric elements are connected, and \textbf{local relations}, which provide the constraints needed for reasoning. GeoFocus learns both through a global-to-local training framework. Its VertexLang Topology Perceptor reconstructs diagrams using compact vertex coordinates and connectivity rather than verbose drawing programs. Its Critical Local Perceptor generates targeted Q\&A supervision from recorded geometric properties and descriptions across 13 task categories. By learning these complementary geometric cues during training, GeoFocus improves the evidence available for downstream reasoning without requiring explicit reconstruction or external tools at inference time.

Across three mainstream GPS benchmarks (Geo3K, GeoQA, and FormalGeo7K), reasoning-enhanced GeoFocus improves average accuracy over Group Relative Policy Optimization (GRPO)-only baselines by 5.2 and 4.9 percentage points (pp) on Qwen3-VL-2B and 4B, respectively, with the 2B variant surpassing several larger general-purpose LMMs. Ablations show that both perception modules improve performance individually and yield larger gains when combined. Further comparisons demonstrate that the Critical Local Perceptor outperforms existing perception Q\&A supervision, while VertexLang achieves stronger downstream GPS performance than code-based reconstruction across training scales. GeoFocus also exhibits greater robustness to increasing visual dependence on MathVerse.

Our contributions are summarized as follows:
\begin{itemize}
    \item We propose a global-to-local perception training framework combining efficient global topology learning with targeted supervision of reasoning-critical local relations.
    \item We introduce VertexLang to compactly encode vertex coordinates and connectivity, and a Critical Local Perceptor to generate targeted Q\&A pairs from geometric properties and descriptions.
\end{itemize}

\begin{figure*}[t!]
    \centering
    \includegraphics[width=1\linewidth]{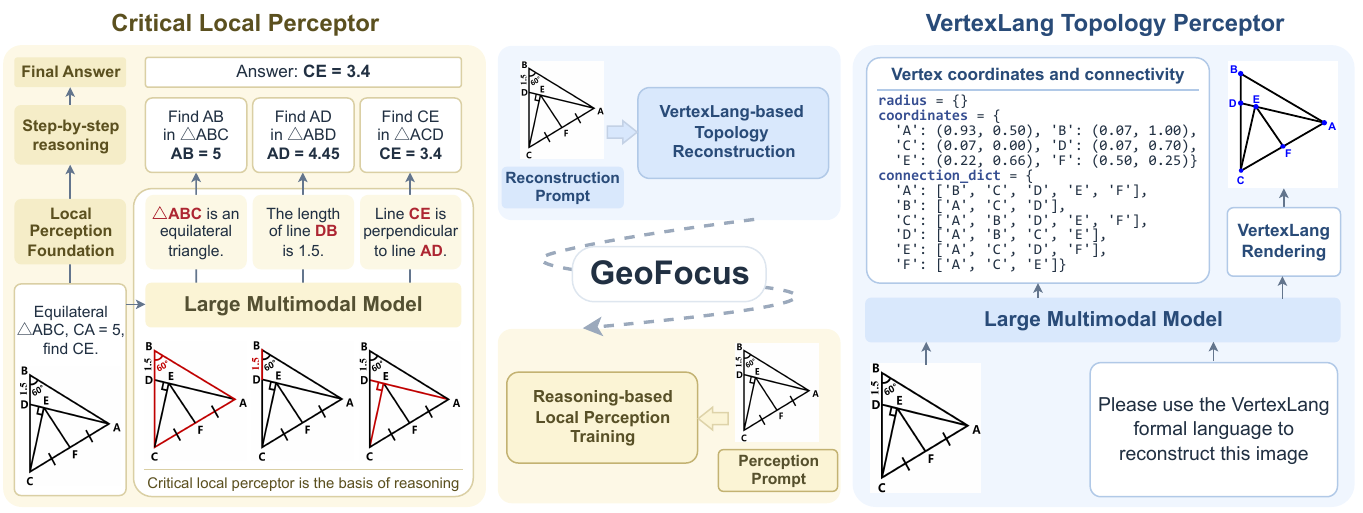}
    \caption{\textbf{Overview of \mymethod}. The Critical Local Perceptor provides targeted perception supervision for reasoning-relevant local geometric relations, while the VertexLang Topology Perceptor provides compact reconstruction supervision for global topology structure.}
    \label{fig:GeoTAR}
\end{figure*}

\section{Related Work} 

GPS perception research spans formal language-based methods, image reconstruction, and perception Q\&A. The first supports symbolic reasoning, while the latter two enhance LMM perception.

\paragraph{Formal Language-Based Methods}
Early GPS methods typically convert geometric diagrams into symbolic representations before reasoning. Rule-based approaches~\cite{lu2021inter,peng2023geodrl,wu2024gps,seo2015solving} use handcrafted parsers and predefined rules for symbolic reasoning. To reduce reliance on handcrafted rules, neural approaches~\cite{li2024lans, zhang2023multi, chen2021geoqa, chen2022unigeo} learn structural and semantic representations with specialized reasoning modules. Recent methods~\cite{wang2026concise,zhang2025geosdf, xia2024geox} further improve flexibility using LMMs or more general symbolic languages, while accurate diagram-to-symbol conversion remains a key bottleneck.
Building on formal geometric representations, GeoFocus instead uses a compact topology language as training supervision rather than an inference-time symbolic interface.

\paragraph{Image Reconstruction Methods}
Inference-time reconstruction methods~\cite{zhou2024image,hu2024visual,meng2023chain} use external tools to reconstruct diagrams, improving spatial reasoning but introducing inference overhead.
Training-based approaches instead incorporate reconstruction objectives directly~\cite{wang2025mathcoder,cheng2025geouni,wei2024slow}. Pixel-level reconstruction~\cite{cheng2025geouni} and segment decomposition~\cite{wei2024slow} emphasize geometric coordinates but capture topology less effectively. Code-based representations~\cite{wang2025mathcoder} better preserve structural relations but require lengthy rendering programs.
Building on reconstruction-based training, VertexLang directly encodes vertex coordinates and connectivity, avoiding verbose rendering procedures.

\paragraph{Perception Q\&A Methods}
Perception Q\&A training improves diagram understanding through specialized supervision. LMM-based methods expand existing geometry data through chain-of-thought (CoT) augmentation~\cite{luo2025ursa,gao2025g}, question synthesis~\cite{shi2024math,peng2024multimath}, and tool-assisted generation~\cite{cai2024geogpt4v}, but may overlook critical local relations or introduce geometric inaccuracies.
In contrast, template-based methods~\cite{huang2025vision,zhang2024mavis,kazemi2023geomverse,pan2025enhancing,fu2025trustgeogen} provide accurate geometric supervision but may offer limited coverage of reasoning-relevant local relations. Building on this strategy, our Critical Local Perceptor explicitly targets 13 categories of reasoning-relevant local relations using geometric properties and descriptions recorded during synthesis.

Prior studies have explored structural representations and perception supervision to enhance geometric perception.
GeoFocus builds on both directions by combining compact topology reconstruction with reasoning-relevant local perception supervision, without additional inference steps.

\section{GeoFocus: Geometric Perception Using Global-to-Local Supervision}
\setlength{\abovedisplayskip}{6pt}
\setlength{\belowdisplayskip}{6pt}
\setlength{\abovedisplayshortskip}{4pt}
\setlength{\belowdisplayshortskip}{4pt}

To address perceptual limitations in GPS, we propose \mymethod{} with two modules (Fig.~\ref{fig:GeoTAR}). The Critical Local Perceptor learns reasoning-relevant local structures from synthetic Q\&A pairs, while the VertexLang Topology Perceptor learns global topology through symbolic image reconstruction.

\subsection{Critical Local Perceptor}
\label{Sec:Axiom}

Motivated by cognitive studies showing that geometric experts recognize local diagram configurations to retrieve relevant knowledge~\cite{koedinger1990abstract}, we introduce the Critical Local Perceptor to explicitly supervise reasoning-relevant local geometric relations. Specifically, two transformation modules leverage reasoning-grounded template pools to generate targeted perception Q\&A pairs from geometric properties and descriptions recorded during diagram synthesis.

\paragraph{Template Design} 
Drawing on geometric properties and relations from existing formal frameworks~\cite{lu2021inter,zhang2024formalgeo}, supplemented by consultations with mathematics teachers, we identify 13 categories of critical local perception tasks for GPS. We assess their coverage across 5,000 middle-school geometry problems, finding that they capture a broad range of local perception requirements. 
These categories fall into two groups: \textit{Basic Measurement}, which concerns quantities such as angles, lengths, and areas; \textit{Relational Reasoning}, which focuses on spatial relations among geometric elements. Detailed category definitions and coverage analysis are provided in the Appendix.

To translate these task categories into targeted perception supervision, we construct four Q\&A template sub-pools, each focusing on a distinct aspect of local geometry. For \textit{Basic Measurement}, the Quantification pool ($W_Q$) queries angle and length measurements, while the Comparison pool ($W_C$) examines relative magnitudes of angles, lengths, and areas. For \textit{Relational Reasoning}, the Line-based Relation pool ($W_L$) targets geometric configurations and structural relations, such as shape properties, angle bisectors, parallelism, and perpendicularity, while the Point-based Relation pool ($W_P$) focuses on positional relations, such as midpoints, points on circles, perpendicular feet, and collinearity. Representative templates are provided in the Appendix.

\begin{figure}[t!]
    \centering
    \includegraphics[width=1\linewidth]{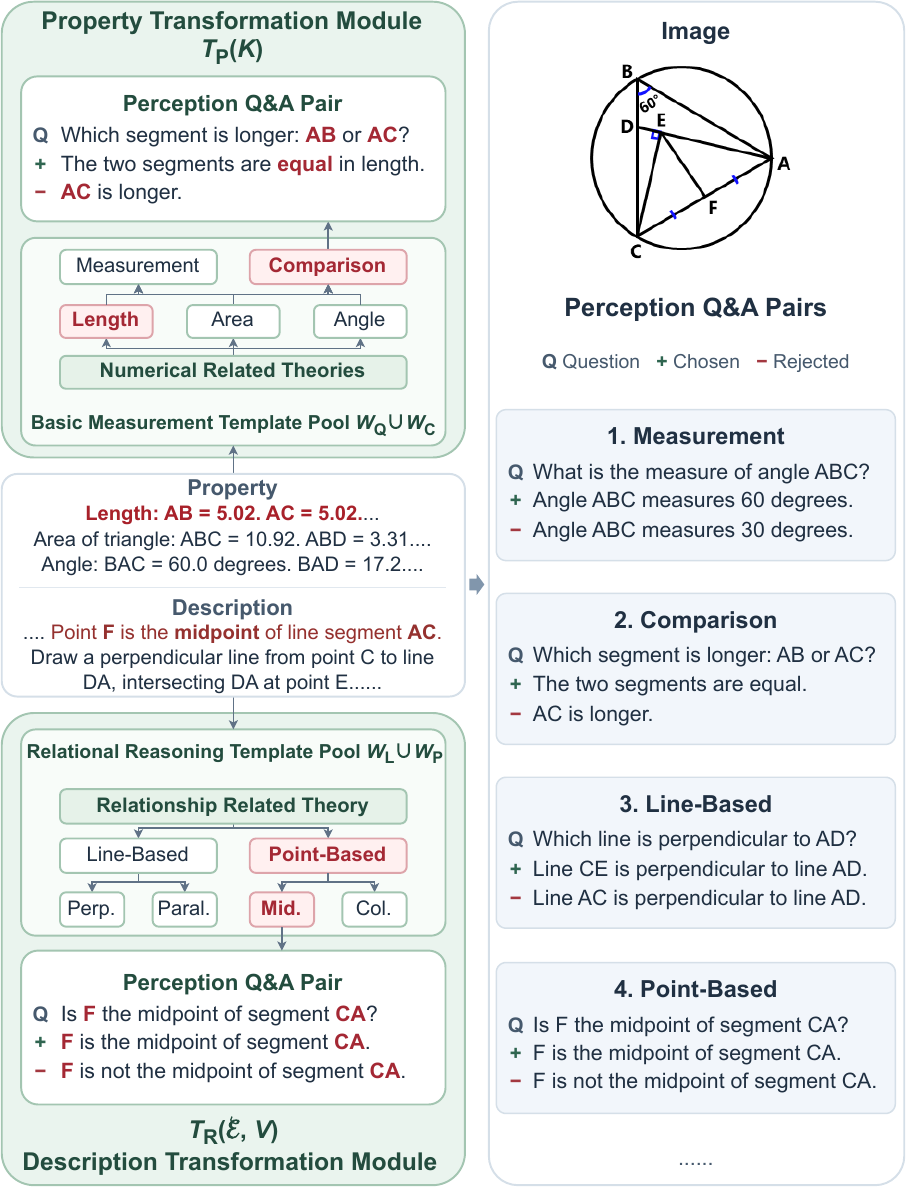}
    \caption{\textbf{Working mechanism of Critical Local Perceptor}. The Property Transformation Module generates Basic Measurement Q\&A pairs from recorded geometric properties, while the Description Transformation Module generates Relational Reasoning Q\&A pairs from recorded descriptions. `Perp.', `Paral.', `Mid.', and `Col.' denote perpendicular, parallel, midpoint, and collinear, respectively.}
    \label{fig:align_engine}
\end{figure}

\paragraph{Perceptor Design} Building on the four template sub-pools, we design the Critical Local Perceptor to convert geometric descriptions and properties into targeted perception Q\&A pairs (Fig.~\ref{fig:align_engine}). 
Following prior work~\cite{zhang2024mavis,kazemi2023geomverse,linger2025theorem}, we progressively record these descriptions and properties during diagram synthesis.
We further design the Property Transformation Module and the Description Transformation Module to instantiate matching Q\&A templates using the recorded properties and descriptions, respectively.

Specifically, the Property Transformation Module $T_P(K)$ converts geometric properties $K$ into numerical perception Q\&A pairs by matching templates from the Basic Measurement Template Pool:
\begin{align}
T_P(K) = H_P\big(W_Q(K) \cup W_C(K_i, K_j)\big).
\end{align}
Here, $H_P$ transforms recorded geometric properties into perception Q\&A pairs through hierarchical template matching. As shown in Fig.~\ref{fig:align_engine}, for the property `Length: AB = 5.02. AC = 5.02', \(H_P\) first matches the length category based on the keyword `Length', then matches the Comparison category based on the inclusion of multiple objects, and finally matches the equal template $TC_{K_i = K_j}$ based on the value to obtain the corresponding Q\&A pair.

The Description Transformation Module $T_R(\mathcal{E}, V)$ generates relational perception Q\&A pairs by matching recorded geometric descriptions with templates from the Line-based Relation ($W_L$) and Point-based Relation ($W_P$) pools, where $\mathcal{E}$ and $V$ denote the sets of geometric lines and points, respectively. Specifically, the transformation function $H_R$ selects and instantiates matching templates based on the entity types and semantic relations in each description:
\begin{align}
T_R(\mathcal{E}, V) = H_R\big(W_L(\mathcal{E}_i, \mathcal{E}_j) \cup W_P(V_i, \mathcal{E}_j)\big).
\end{align}
As illustrated in Fig.~\ref{fig:align_engine}, given the description `Point F is the midpoint of line segment AC', $H_R$ first identifies $F$ as a point and $AC$ as a segment, then uses the `Midpoint' predicate to select and instantiate the corresponding template from $W_P$. Examples covering all 13 task categories are provided in the Appendix.

\subsection{VertexLang Topology Perceptor}
\label{Sec:VertexLang}

Accurate perception of local geometric relations relies on identifying the relevant geometric elements and their spatial configurations, motivating complementary supervision through global topology reconstruction. To improve reconstruction efficiency, we introduce \textbf{VertexLang}, a compact formal representation that directly encodes vertex coordinates and connectivity.

\begin{figure}[t!]
    \centering
    \includegraphics[width=1\linewidth]{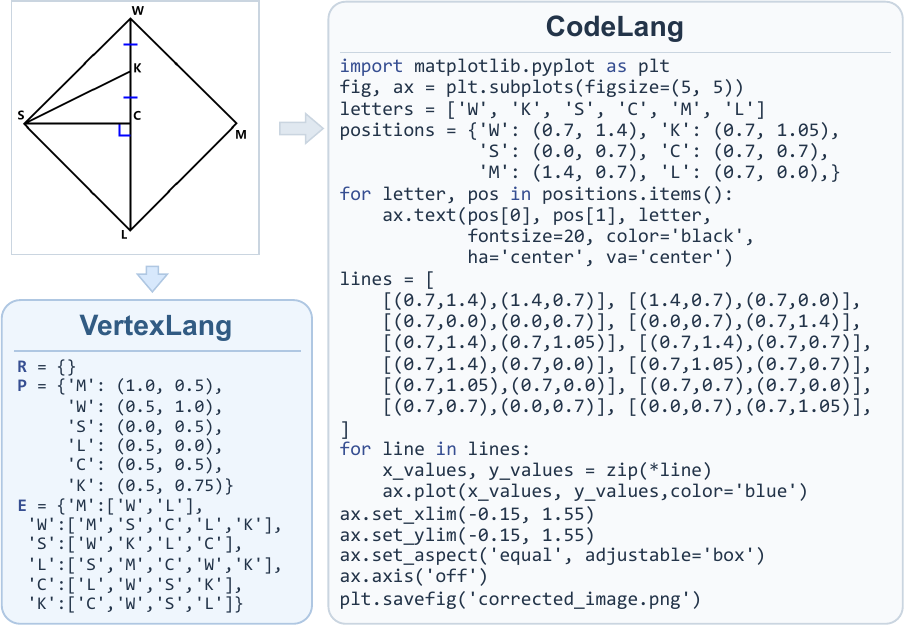}
    \caption{Qualitative comparison of VertexLang-based and CodeLang-based image reconstruction.}
    \label{fig:recon_number}
\end{figure}

\begin{figure*}[t!]
    \centering
    \includegraphics[width=1\linewidth]{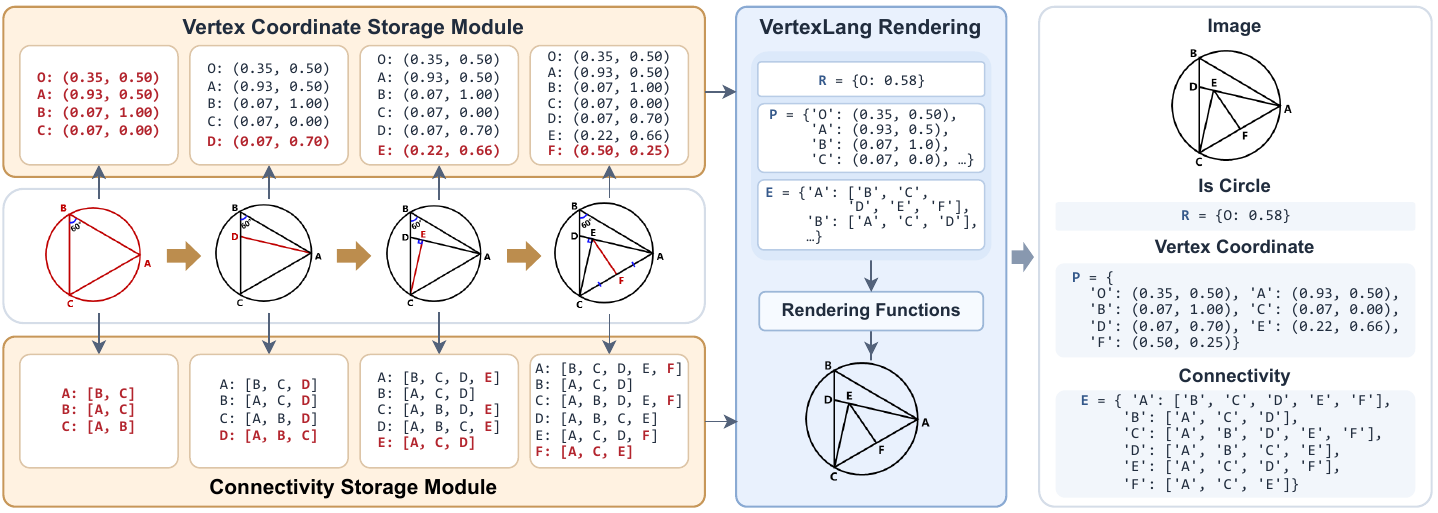}
    \caption{\textbf{Working Mechanism of VertexLang Reconstructor}. The circle-radius dictionary $\mathbf{R}$ stores circle radii from the geometry substrate. As elements are added, the Vertex Coordinate Storage Module updates $\mathbf{P}$ with vertex coordinates, while the Connectivity Storage Module updates $\mathbf{E}$ with neighbor sets. The VertexLang Rendering Module then renders the diagram from the three dictionaries.}
    \label{fig:recon_engine}
\end{figure*}

\paragraph{Definition of VertexLang}
Specifically, VertexLang represents geometric diagrams using three dictionaries: a circle radius dictionary $\mathbf{R}$, a vertex coordinate dictionary $\mathbf{P}$, and a connectivity dictionary $\mathbf{E}$, without encoding rendering procedures.
The radius dictionary $\mathbf{R}$ records the radii of circles in geometric diagrams:
\begin{align}
\mathbf{R} = \left\{o_i : r_i | o_i \in \mathcal{O}, r_i > 0\right\},
\end{align}
where $o_i:r_i$ denotes a center-radius key-value pair, with $o_i$ representing the center label and $r_i$ its radius. Here, $\mathcal{O} \subseteq V$ denotes the set of circle centers within the vertex set $V$. For diagrams without circles, $\mathbf{R}$ is empty. 
The vertex-coordinate dictionary $\mathbf{P}$ maps each vertex to its normalized coordinates:
\begin{align}
\mathbf{P}=\left\{v_i:(x_i, y_i)|v_i \in V, x_i \in [0,1], y_i \in [0,1]\right\},
\end{align}
where $v_i$ denotes a vertex label and $(x_i,y_i)$ its normalized coordinates. Vertex coordinates and circle radii are normalized using a common scale to preserve relative geometric proportions and encourage the model to learn spatial relationships among vertices rather than rely on absolute coordinate magnitudes.
Finally, the connectivity dictionary $\mathbf{E}$ represents straight-line connections between vertices:
\begin{equation}
\begin{aligned}
    \mathbf{E} &= \{v_i \mapsto \mathcal{N}(v_i) \mid v_i \in V\},&\mathcal{N} : V \rightarrow 2^V,
\end{aligned}
\end{equation}
\begin{equation}
\begin{aligned}
    \mathcal{N}(v_i) &= \{v_j \in V \mid v_j \sim v_i\},
\end{aligned}
\end{equation}
where $\mathcal{N}(v_i)$ denotes the neighbors of vertex $v_i$, $2^V$ is the power set of $V$, and $v_j \sim v_i$ indicates a straight-line connection between the two vertices.
During diagram rendering, a predefined rendering function $\mathcal{F}$ maps the three dictionaries to the corresponding geometric image $G$:
\begin{align}
G=\mathcal{F}(\mathbf{R, P, E}).
\end{align}

To evaluate the efficiency of compact geometric representation, we compare VertexLang with code-based reconstruction (CodeLang)~\cite{wang2025mathcoder} on 2,500 geometric diagrams of varying complexity. We use Gemini 2.0 Flash to generate the corresponding reconstruction code. On the same diagrams, VertexLang reduces serialized annotation length by 88\%, as illustrated in Fig.~\ref{fig:recon_number}. Training with VertexLang also takes 20\% less time than with code-based annotations.

\paragraph{VertexLang Reconstructor}
To generate paired geometric diagrams and VertexLang annotations for training, we develop the VertexLang Reconstructor, as illustrated in Fig.~\ref{fig:recon_engine}. Building on prior incremental diagram synthesis methods~\cite{linger2025theorem,fu2025trustgeogen}, we progressively add geometric elements while additionally recording the VertexLang annotations. The circle-radius dictionary $\mathbf{R}$ is initialized from circular elements when present and remains empty otherwise. As geometric elements are added, the Vertex Coordinate Storage Module records vertex coordinates $(x,y)$ in $\mathbf{P}$, while the Connectivity Storage Module updates the corresponding neighbor sets $\mathcal{N}(v)$ in $\mathbf{E}$. Finally, the VertexLang Rendering Module generates the geometry image from the three dictionaries, yielding paired images and VertexLang annotations. Detailed pseudocode is provided in the Appendix.

\section{Experiments}

\subsection{Setup}

\paragraph{Implementation Details}
We evaluate GeoFocus on Qwen3-VL-2B/4B/8B~\cite{bai2025qwen3}, Qwen2.5-VL-3B/7B~\cite{bai2025qwen2}, and Gemma3-4B~\cite{team2025gemma}. We conduct module-level ablations on Qwen3-VL to examine the contributions of global and local perception across model scales while investigating design choices and generalization on Qwen2.5-VL. 
Stage 1 trains the VertexLang Topology Perceptor with DynamicGT-RL on 15k synthetic samples for 15 epochs at a learning rate of $2\times10^{-6}$, using EasyR1~\cite{zheng2025easyr1} with default settings otherwise. Stage 2 trains the Critical Local Perceptor with DPO on 80k synthetic perception Q\&A pairs for 1 epoch at a learning rate of $5\times10^{-6}$, using LlamaFactory~\cite{zheng2024llamafactory}.
Both stages use 8 GPUs, requiring 10.5 and 4 hours, respectively. GeoFocus introduces no additional inference overhead, external tools, or test-time scaling. Benchmark details and evaluation protocols are provided in the Appendix.

\paragraph{Training Strategy}
For the Critical Local Perceptor, we adopt Direct Preference Optimization (DPO) to accommodate diverse answer formats and mitigate overfitting to non-unique token expressions. Given a query $Q$, image $I$, and positive/negative responses $R^+$/$R^-$, the objective is
\begin{multline}
\mathcal{L}_{\text{DPO}}
= -\mathbb{E}_{\mathbf{Q}, \mathbf{I}, \mathbf{R}^+, \mathbf{R}^-}
\Bigg[ 
\log \sigma 
\Bigg( 
\beta \log \frac{P_\theta(\mathbf{R}^+|\mathbf{Q},\mathbf{I})}
{P_{\text{ref}}(\mathbf{R}^+|\mathbf{Q},\mathbf{I})} \\
- \beta \log \frac{P_\theta(\mathbf{R}^-|\mathbf{Q},\mathbf{I})}
{P_{\text{ref}}(\mathbf{R}^-|\mathbf{Q},\mathbf{I})}
\Bigg)
\Bigg],
\end{multline}
This preference-based supervision encourages correct responses while suppressing incorrect ones, providing robust perceptual priors for subsequent reasoning.

When training the VertexLang Topology Perceptor, the model initially struggles to produce accurate structured reconstructions, making reinforcement learning difficult to optimize~\cite{yue2026does}. We therefore introduce \textbf{DynamicGT-RL}, which probabilistically incorporates the ground-truth (GT) samples into advantage estimation:
\begin{equation}
\hat{A}_i(R(P))= 
\begin{cases}
\frac{R\left(p_i\right)-\text { mean }\left(R\left(P_{\text {model }}\right)\right)}{\operatorname{std}\left(R\left(P_{\text {model }}\right)\right)}, & x \geq \tau, \\ 
\frac{R\left(p_i\right)-\operatorname{mean}\left(R\left(P_{\text {model }} \cup P_{\text {gt }}\right)\right)}{\operatorname{std}\left(R\left(P_{\text {model }} \cup P_{\text {gt }}\right)\right)}, & x<\tau,
\end{cases}
\label{eq:ADV}
\end{equation}
where $x\sim U(0,1)$ and $\tau$ controls the GT inclusion ratio.

For VertexLang reconstruction, we define
\begin{equation}
R_{\text{recon}}
=0.5R_{\text{coord}}+0.5R_{\text{conn}},
\end{equation}
where $R_{\text{coord}}$ measures vertex-coordinate accuracy based on Euclidean distance, and $R_{\text{conn}}$ measures edge connectivity using F1 score. We use $R_{\text{recon}}$ as $R$ in Eq.~\ref{eq:ADV}. More implementation details are provided in the Appendix.

\subsection{Ablation Study}

We first evaluate the contributions of the two perception modules to GPS (Sec.~\ref{sec:effectiveness-components}), followed by an analysis of the design choices in the Critical Local Perceptor (Sec.~\ref{sec:Axiom Ablation}) and VertexLang Topology Perceptor (Sec.~\ref{sec:Topo Ablation}). The Appendix further examines training strategies and module ordering, and reports consistent gains on Gemma3-4B.

\begin{table}[t]
\centering
\caption{Ablation of \mymethod{} on Qwen3-VL with and without reasoning enhancement.
`Topo' and `Local' denote the VertexLang Topology Perceptor and Critical Local Perceptor, respectively, both trained solely on synthetic data. 
The reasoning-enhanced group additionally uses GRPO for geometric reasoning training.
Total denotes the sum of Top-1 accuracy across the three benchmarks.}
\label{stage}

\footnotesize
\setlength{\tabcolsep}{3.5pt}
\renewcommand{\arraystretch}{0.8}

\resizebox{\columnwidth}{!}{\begin{tabular}{@{}c cc ccc c@{}}
\toprule
Size & Topo & Local & Geo3K & GeoQA & FormalGeo7K & Total \\
\midrule

\multicolumn{7}{c}{\textit{General LMMs}} \\
\midrule

\multirow{4}{*}{2B}
&  &   & 29.5 & 42.1 & 33.5 & 105.1 \\  & \ding{51} &   & 31.5 & 42.3 & 34.3 & 108.1 {\scriptsize($\uparrow$3.0)} \\
&   & \ding{51} & 34.1 & 44.6 & 34.5 & 113.2 {\scriptsize($\uparrow$8.1)} \\
& \ding{51} & \ding{51} & 34.8 & 46.7 & 36.3 & \textbf{117.8} {\scriptsize($\uparrow$12.7)} \\
\midrule

\multirow{4}{*}{4B}
&   &   & 49.7 & 63.0 & 55.9 & 168.6 \\
& \ding{51} &   & 50.8 & 64.3 & 56.3 & 171.4 {\scriptsize($\uparrow$2.8)} \\
&   & \ding{51} & 54.3 & 65.3 & 58.7 & 178.3 {\scriptsize($\uparrow$9.7)} \\
& \ding{51} & \ding{51} & 54.4 & 66.6 & 60.5 & \textbf{181.5} {\scriptsize($\uparrow$12.9)} \\
\midrule

\multicolumn{7}{c}{\textit{Reasoning-Enhanced LMMs}} \\
\midrule

\multirow{4}{*}{2B}
&   &   & 37.9 & 51.9 & 42.0 & 131.8 \\
& \ding{51} &   & 38.6 & 52.6 & 42.6 & 133.8 {\scriptsize($\uparrow$2.0)} \\
&   & \ding{51} & 41.1 & 54.6 & 45.1 & 140.8 {\scriptsize($\uparrow$9.0)} \\
& \ding{51} & \ding{51} & 42.8 & 56.9 & 47.8 & \textbf{147.5} {\scriptsize($\uparrow$15.7)} \\
\midrule

\multirow{4}{*}{4B}
&   &   & 58.1 & 70.8 & 64.4 & 193.3 \\
& \ding{51} &   & 58.6 & 72.2 & 65.0 & 195.8 {\scriptsize($\uparrow$2.5)} \\
&   & \ding{51} & 61.2 & 73.7 & 66.7 & 201.6 {\scriptsize($\uparrow$8.3)} \\
& \ding{51} & \ding{51} & 65.1 & 74.4 & 68.6 & \textbf{208.1} {\scriptsize($\uparrow$14.8)} \\

\bottomrule
\end{tabular}
}
\end{table}

\subsubsection{Ablation of the Proposed Modules} 
\label{sec:effectiveness-components}

We evaluate the individual and combined effects of the VertexLang Topology Perceptor and Critical Local Perceptor on Qwen3-VL-2B/4B. As shown in the General LMMs section of Tab.~\ref{stage}, both modules consistently improve GPS performance across all three benchmarks. The VertexLang Topology Perceptor independently improves the summed accuracy by 3.0 and 2.8 percentage points (pp) on the 2B and 4B models, while the Critical Local Perceptor achieves gains of 8.1 and 9.7 pp. Combining both yields larger gains of 12.7/12.9 pp, demonstrating their complementary benefits for GPS.

To assess whether perception improvements persist under stronger reasoning capabilities, we further apply outcome-supervised GRPO~\cite{shao2024deepseekmath} to enhance geometric reasoning, using the training splits of Geo3K, GeoQA, and FormalGeo7K while strictly holding out evaluation sets. We exclude the CoT-augmented Geo170K~\cite{gao2025g} to isolate the benefits of perception training from additional reasoning supervision. As shown in the Reasoning-Enhanced LMMs section of Tab.~\ref{stage}, GeoFocus outperforms GRPO-only baselines by 15.7/14.8 pp in summed accuracy on 2B/4B models. These results demonstrate that perception supervision remains effective and complementary to reasoning optimization.

\begin{table}[t]
\centering
\caption{Comparison with CogAlign on the CogAlign-data perception test set. `Local' denotes our Critical Local Perceptor. `Perp.' denotes Perpendicular.}
\label{perception_compare}

\footnotesize
\setlength{\tabcolsep}{2.2pt}
\renewcommand{\arraystretch}{0.9}

\resizebox{\columnwidth}{!}{\begin{tabular}{@{}c l c c c c c@{}}
\toprule
Size & Config
& \makecell{Angle\\Comparison}
& \makecell{Perp.\\Detection}
& \makecell{Parallel\\Comparison}
& \makecell{Chart\\Projection}
& Avg. \\
\midrule

\multirow{3}{*}{3B}
& Baseline
& 62.5 & 54.1 & 57.0 & 79.5 & 63.3 \\
& CogAlign
& 70.1 & 55.7 & 49.6 & 83.4 & 64.7 \\
& Local
& \textbf{91.6} & \textbf{55.7} & \textbf{60.7} & 81.9 & \textbf{72.5} \\
\midrule

\multirow{3}{*}{7B}
& Baseline
& 87.1 & 59.0 & 63.7 & 90.2 & 75.0 \\
& CogAlign
& 84.6 & 60.1 & 68.7 & 90.2 & 75.9 \\
& Local
& \textbf{89.1} & \textbf{60.8} & \textbf{77.1} & \textbf{90.5} & \textbf{79.4} \\
\bottomrule
\end{tabular}
}
\end{table}

\begin{figure}[t!]
    \centering
    \includegraphics[width=1\linewidth]{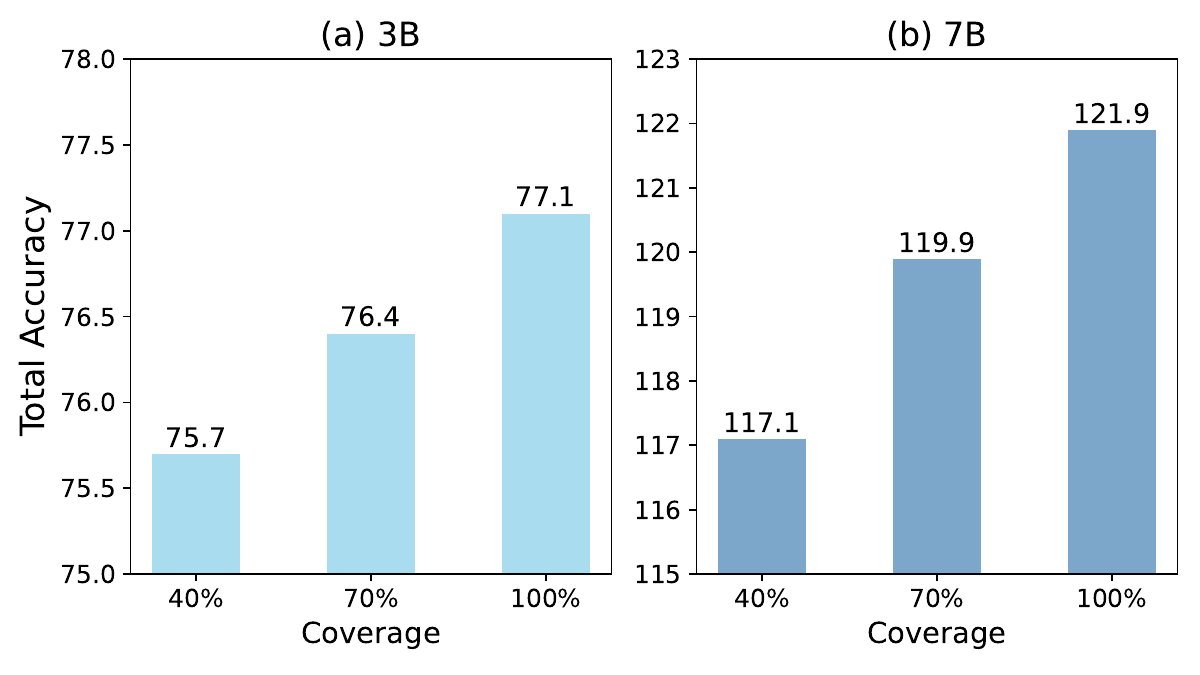}
    \caption{\textbf{Impact of critical local task-category coverage on GPS performance.} (a) Qwen2.5-VL-3B and (b) Qwen2.5-VL-7B.
Coverage denotes the proportion of the 13 local perception task categories included in training.}
    \label{fig:coverage}
\end{figure}

\subsubsection{Ablation of Critical Local Perceptor Design}
\label{sec:Axiom Ablation}

We evaluate the Critical Local Perceptor by examining the impact of local task-category coverage and comparing it with existing perception supervision methods. Experiments on Qwen2.5-VL-3B/7B further assess the effectiveness of our design across model scales.

\paragraph{Effect of Critical Local Task-Category Coverage}
To assess the effect of local task-category coverage, we train models using perception Q\&A data spanning approximately 40\%, 70\%, and 100\% of the 13 predefined task categories, while keeping the total number of training samples and optimization steps fixed. As shown in Fig.~\ref{fig:coverage}, GPS performance consistently increases with category coverage, suggesting that broader supervision across these local perception tasks benefits geometric reasoning.

\paragraph{Advantage of Critical Local Perceptor in Improving Perception Performance} For a fair comparison, we train Critical Local Perceptor and CogAlign with 65k samples each, rather than the full dataset used in the coverage ablation. As shown in Tab.~\ref{perception_compare}, the 3B and 7B models trained with the Critical Local Perceptor achieve an average improvement of 5.7 pp on the CogAlign-data test set compared to those trained on CogAlign’s native samples.
We attribute this improvement to the broader coverage of local geometric structures. Compared with CogAlign's eight perception tasks, Critical Local Perceptor covers thirteen task categories, providing more comprehensive supervision for reasoning-relevant local perception. See the Appendix for task-type comparisons.

\paragraph{Advantage of Critical Local Perceptor in Enhancing Reasoning Performance}
As shown in Tab.~\ref{perception_data}, 
CogAlign slightly reduces GPS accuracy by 0.1/2.0 pp on 3B/7B, whereas our Critical Local Perceptor improves it by 1.4/2.0 pp. Combining both methods yields gains of 4.0/2.6 pp over the baseline, demonstrating their complementary benefits for GPS.

\begin{table}[t]
\centering
\caption{Comparison with the synthetic geometry perception Q\&A method CogAlign on real-world GPS datasets. `Local' denotes our Critical Local Perceptor.}
\label{perception_data}

\footnotesize
\setlength{\tabcolsep}{3pt}
\renewcommand{\arraystretch}{0.9}

\resizebox{\columnwidth}{!}{\begin{tabular}{@{}c l c c c c@{}}
\toprule
Size & Config & Geo3K & GeoQA & FormalGeo7K & Total \\
\midrule

\multirow{4}{*}{3B}
& Baseline        & 25.3 & 32.1 & 16.8 & 74.2 \\
& CogAlign        & 25.5 & 31.6 & 17.0 & 74.1 {\scriptsize($\downarrow$0.1)} \\
& Local           & 25.8 & 32.6 & 17.2 & 75.6 {\scriptsize($\uparrow$1.4)} \\
& CogAlign + Local
                  & 26.5 & 34.9 & 16.8 & \textbf{78.2} {\scriptsize($\uparrow$4.0)} \\
\midrule

\multirow{4}{*}{7B}
& Baseline        & 39.4 & 44.4 & 35.6 & 119.4 \\
& CogAlign        & 38.1 & 43.4 & 35.9 & 117.4 {\scriptsize($\downarrow$2.0)} \\
& Local           & 39.1 & 46.3 & 36.0 & 121.4 {\scriptsize($\uparrow$2.0)} \\
& CogAlign + Local
                  & 40.3 & 45.0 & 36.7 & \textbf{122.0} {\scriptsize($\uparrow$2.6)} \\
\bottomrule
\end{tabular}
}
\end{table}

\subsubsection{Ablation of VertexLang Topology Perceptor Design}
\label{sec:Topo Ablation}
We conduct ablations on Qwen2.5-VL-3B/7B to evaluate the advantages of VertexLang Topology Perceptor over existing methods and its design choices.

\begin{figure}[t!]
    \centering
    \includegraphics[width=0.95\linewidth]{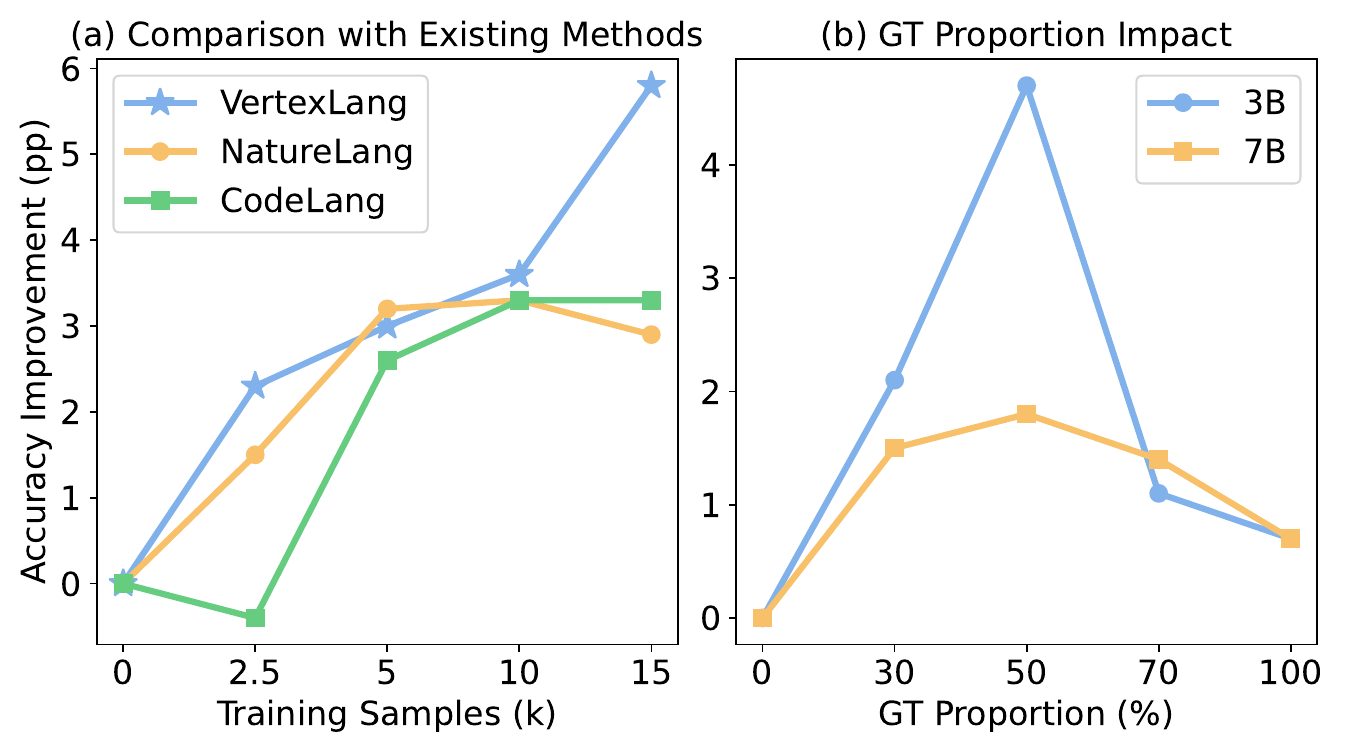}
    \caption{(a) Comparison with existing methods at different sample scales. (b) Impact of the ground-truth (GT) inclusion ratio in DynamicGT-RL. The vertical axis shows the summed accuracy improvement for Geo3K, GeoQA, and FormalGeo7K.}
    \label{fig:combine}
\end{figure}

\paragraph{Advantage of VertexLang at Different Sample Scales}
We compare three global topology perception methods across sample scales: TR-CoT captioning (NatureLang)~\cite{linger2025theorem}, code-based reconstruction (CodeLang)~\cite{wang2025mathcoder}, and VertexLang reconstruction. For fairness, all methods use Qwen2.5-VL-3B trained with DynamicGT-RL, a 50\% GT proportion, and edit-distance rewards. 
As shown in Fig.~\ref{fig:combine}(a), VertexLang outperforms CodeLang at all scales and NatureLang at most scales. Moreover, VertexLang continues improving up to 15k samples, whereas NatureLang and CodeLang plateau or degrade beyond 10k. These results demonstrate that compact reconstruction supervision improves training efficiency and downstream GPS performance.

We conduct two additional experiments: evaluating reconstruction quality using coordinate and connectivity accuracy, and comparing downstream GPS performance between compact VertexLang and its complete image-rendering code as reconstruction targets. VertexLang performs better in both evaluations, with details provided in the Appendix.

\begin{table}[t]
\centering
\caption{Comparison of out-of-domain generalization.}
\label{out-of-domain}

\footnotesize
\setlength{\tabcolsep}{3.3pt}
\renewcommand{\arraystretch}{0.9}

\resizebox{\columnwidth}{!}{\begin{tabular}{@{}c l c c c c c c c@{}}
\toprule
Size & Model
& \makecell{Hallu\\Bench}
& \makecell{We\\Math}
& \makecell{Math\\Verse}
& \makecell{Math\\Vision}
& \makecell{Math\\Vista}
& \makecell{Chart\\QA}
& Total \\
\midrule

\multirow{4}{*}{3B}
& Baseline
& 60.5 & 51.7 & 34.3 & 21.3 & 56.4 & 73.2 & 297.4 \\
& SFT
& 57.8 & 11.9 & 25.2 & 19.0 & 51.5 & 70.3 & 235.7 \\
& GRPO
& 63.3 & 61.6 & 39.3 & 24.3 & 62.1 & 76.4 & 327.0 \\
& \mymethod
& \textbf{64.7} & \textbf{62.5} & \textbf{39.4}
& \textbf{24.5} & \textbf{63.9} & \textbf{76.6}
& \textbf{331.6} \\
\midrule

\multirow{4}{*}{7B}
& Baseline
& 68.0 & 62.9 & 43.1 & 25.3 & 68.2 & 79.0 & 346.5 \\
& SFT
& 57.5 & 20.2 & 15.4 & 18.5 & 55.6 & 71.1 & 238.3 \\
& GRPO
& 68.7 & 68.9 & 45.7 & 27.7 & 72.1 & 81.3 & 364.4 \\
& \mymethod
& \textbf{71.1} & \textbf{69.4} & \textbf{45.7}
& \textbf{28.0} & \textbf{74.3} & \textbf{81.5}
& \textbf{370.0} \\
\bottomrule
\end{tabular}
}
\end{table}

\paragraph{Ablation of GT Ratio in DynamicGT-RL}
We ablate the GT addition ratio in DynamicGT-RL. As shown in Fig.~\ref{fig:combine}(b), incorporating GT improves performance, with 50\% achieving the best results. Higher ratios lead to diminishing performance, which we hypothesize may stem from excessive GT guidance limiting generalization.

\subsection{Generalization and Visual Robustness}

\paragraph{Out-of-Domain Generalization Ability}
We evaluate \mymethod{} on additional multimodal benchmarks covering mathematical reasoning, chart understanding, and hallucination detection, using Qwen2.5-VL-3B/7B (Tab.~\ref{out-of-domain}). While geometry-specific Supervised Fine-Tuning (SFT) substantially degrades performance on several benchmarks, GRPO partially recovers this loss. In contrast, \mymethod{} consistently matches or outperforms GRPO, achieving an average gain of 4.8 pp across six benchmarks and two model scales over the original Qwen2.5-VL backbones. Notably, the improvements it achieves on HalluBench indicate lower sensitivity to misleading visual cues.

\begin{table}[t]
\centering
\caption{Human evaluation of geometric perception errors on 200 sampled problems.
`Topo' and `Local' denote the VertexLang Topology Perceptor and Critical Local Perceptor, respectively.
Global and local errors are annotated separately and may co-occur. Error Cases counts problems containing at least one perception error, while Total Errors sums the global and local error counts.}
\label{tab:perception_error}

\footnotesize
\setlength{\tabcolsep}{3.5pt}
\renewcommand{\arraystretch}{0.85}

\resizebox{\columnwidth}{!}{\begin{tabular}{@{}cc cccc@{}}
\toprule
Topo & Local & Global Errors & Local Errors & Error Cases & Total Errors \\
\midrule

     &      & 94 & 49 & 118 & 143 \\
\ding{51} &      & 71 & 47 & 101 & 118 \\
     & \ding{51} & 75 & 42 & 101 & 117 \\
\ding{51} & \ding{51} & \textbf{67} & \textbf{35} & \textbf{80} & \textbf{102} \\

\bottomrule
\end{tabular}
}
\end{table}

\begin{figure}[t!]
    \centering
    \includegraphics[width=1\linewidth]{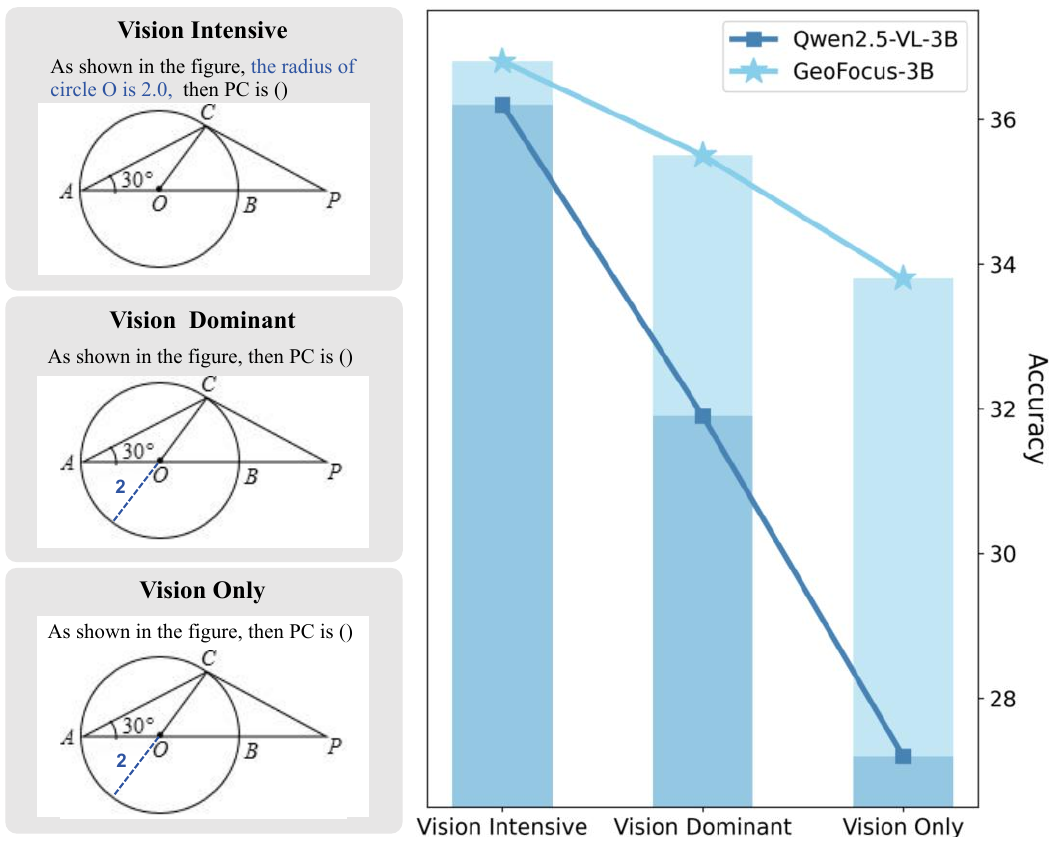}
    \caption{MathVerse accuracy under increasing visual dependence, from Vision Intensive to Vision Dominant to Vision Only.}
    \label{fig:puma}
\end{figure}

\paragraph{Reducing Visual Perception Limitations}
\label{Discussion}
We further assess geometric perception through visual robustness and human error analysis. Additional topology reconstruction results and qualitative examples are provided in the Appendix.

We evaluate visual robustness under the MathVerse protocol~\cite{zhang2024mathverse} across three settings with increasing visual dependence (Fig.~\ref{fig:puma}). \mymethod{} consistently outperforms the baseline, with smaller performance drops as textual cues decrease. This suggests improved robustness when problem-relevant information must be extracted directly from diagrams.

To further assess geometric perception, we manually sampled 200 problems proportionally from Geo3K (50), GeoQA (63), and FormalGeo7K (87).
We examine Qwen3-VL-4B outputs under four training configurations, separately identifying global topology errors (e.g., incorrect connectivity) and local perception errors (e.g., misidentified geometric properties or relations), including those in correctly answered problems.
As shown in Tab.~\ref{tab:perception_error}, Topo reduces global errors from 94 to 71, while Local reduces local errors from 49 to 42.
Combining both further reduces these errors to 67 and 35, respectively, decreasing the number of problems with perception errors from 118 to 80 (32.2\%).
Because perception errors are annotated even for correctly answered problems, this reduction provides evidence of improved geometric grounding beyond gains in final-answer accuracy.

\begin{table}[t!]
\centering
\caption{Top-1 accuracy (\%) on mainstream geometry problem-solving benchmarks. $^*$ denotes results reported in existing papers.}
\label{sota}
\footnotesize
\setlength{\tabcolsep}{1.2mm}
\renewcommand{\arraystretch}{1}
\resizebox{\columnwidth}{!}{
\begin{tabular}{@{}l|cc|ccc@{}}
\hline
Model & Tools & TTS & Geo3K & GeoQA & FormalGeo7K \\
\hline
\multicolumn{6}{c}{Closed-source LMMs} \\ \hline
Gemini 2.0 Flash~\cite{google2025gemini20flash}
& No & No & 58.2 & 69.1 & 36.5 \\
GPT-4o~\cite{hurst2024gpt}
& No & No & 34.6 & 34.0 & 36.7 \\
Claude 3.7~\cite{anthropic2025claude37}
& No & No & 31.1 & 26.9 & 24.0 \\
Gemini 2.5 Pro~\cite{fu2025trustgeogen,comanici2025gemini}
& No & Native & 80.7* & 79.6* & -- \\
OpenAI o3~\cite{fu2025trustgeogen,openai2025o3}
& No & Native & 81.0* & 79.3* & -- \\

\hline
\multicolumn{6}{c}{Open-source LMMs} \\ \hline
Chameleon-7B~\cite{team2024chameleon}
& No & No & 3.9* & 7.3* & 5.9* \\
mPLUG-Owl2-7B~\cite{ye2024mplug}
& No & No & 12.7 & 13.0 & 6.6 \\
Monkey-Chat-7B~\cite{li2024monkey}
& No & No & 4.7 & 10.7 & 9.3 \\
DeepSeek-VL-7B~\cite{lu2024deepseek}
& No & No & 5.2 & 12.2 & 5.8 \\
InternVL-2.5-8B~\cite{chen2024expanding}
& No & No & 16.6 & 30.9 & 11.8 \\
InternVL-3-2B~\cite{zhu2025internvl3}
& No & No & 31.1 & 31.2 & 17.4 \\
InternVL-3-8B~\cite{zhu2025internvl3}
& No & No & 33.9 & 38.6 & 28.8 \\
Gemma3-4B~\cite{team2025gemma}
& No & No & 30.6 & 42.6 & 37.0 \\
Qwen3-VL-4B~\cite{bai2025qwen3}
& No & No & 49.7 & 63.0 & 55.9 \\
Qwen3-VL-8B~\cite{bai2025qwen3}
& No & No & 55.4 & 69.3 & 61.8 \\

\hline
\multicolumn{6}{c}{Open-source Mathematical Geometry Agents} \\ \hline
ReTTS-PGPS-3B~\cite{kang2026reactivating}
& Python & No & 64.7* & 58.7* & -- \\
ReTTS-PGPS-8B~\cite{kang2026reactivating}
& Python & No & 74.5* & 67.2* & -- \\
MetaReason-8B~\cite{yin2026metareason}
& Sketch & Yes & -- & 87.5* & -- \\
GeoWeave-8B~\cite{dong2026beyond}
& No & Yes & 83.8 & 85.2 & 79.2 \\

\hline
\multicolumn{6}{c}{Open-source Mathematical Geometry LMMs} \\ \hline
G-LLaVA-13B~\cite{gao2025g}
& No & No & 35.0* & 20.1* & 14.1* \\
GeoUni~\cite{cheng2025geouni}
& No & No & 50.0* & 66.7* & 59.8* \\
GeoGen-SFT-3B~\cite{pan2025enhancing}
& No & No & 38.6* & 61.4* & -- \\
GeoGen-SFT-7B~\cite{pan2025enhancing}
& No & No & 46.3* & 64.6* & -- \\
\mymethod-2B
& No & No & 42.8 & 56.9 & 47.8 \\
\mymethod-4B
& No & No & 65.1 & 74.4 & 68.6 \\
\mymethod-8B
& No & No & \textbf{68.1} & \textbf{77.1} & \textbf{71.8} \\
\hline
\end{tabular}
}

\vspace{1mm}
\begin{minipage}{\columnwidth}
\scriptsize
\textit{Notes.} Tools indicates external tools invoked during inference. TTS denotes explicit test-time scaling through multi-candidate sampling, search, or ensembling. Native indicates built-in reasoning capabilities without explicit multi-sample TTS.
\end{minipage}
\end{table}

\subsection{Comparison with Existing Methods}
As shown in Tab.~\ref{sota}, 
GeoFocus-2B outperforms several larger general-purpose LMMs, while the 4B and 8B variants improve over their original backbones by 13.2 and 10.2 pp on average. Without external tools or explicit test-time scaling, GeoFocus-8B achieves competitive accuracy and runs 11.6$\times$ faster than GeoWeave-8B on identical hardware, despite lower accuracy.

\section{Conclusion}
GeoFocus demonstrates that improving what a multimodal model perceives can complement improvements in how it reasons. By jointly learning compact global topology through VertexLang and reasoning-relevant local relations through the Critical Local Perceptor, GeoFocus improves geometry problem-solving without additional inference procedures. Consistent gains over matched reasoning-trained baselines, together with a reduction in manually annotated perception errors, indicate that these benefits extend beyond final-answer accuracy. These findings highlight perception-focused training as a practical direction for strengthening multimodal geometric reasoning.

\section*{LLM Usage Statement}
We used OpenAI ChatGPT for manuscript-level feedback and to assist with language editing, text organization. Gemini 2.0 Flash was used to generate image-reconstruction code for the code-based comparison. Generative AI tools were also used to assist with implementation code, which was reviewed and validated by the authors. Generative AI tools were additionally used to refine the visual presentation of selected figures. All AI-assisted figure modifications were reviewed to ensure that the underlying geometric content and experimental results remained unchanged. The research hypotheses, scientific claims, and conclusions were established and verified by the authors, who take full responsibility for all submitted content.

{
    \small
    \bibliographystyle{ieeenat_fullname}
    \bibliography{main}
}

\clearpage
\appendix

\setlength{\emergencystretch}{1.5em}
\raggedbottom
\setlength{\textfloatsep}{10pt plus 2pt minus 2pt}
\setlength{\floatsep}{8pt plus 2pt minus 2pt}
\setlength{\intextsep}{9pt plus 2pt minus 2pt}
\setlength{\dbltextfloatsep}{12pt plus 2pt minus 2pt}
\setlength{\dblfloatsep}{10pt plus 2pt minus 2pt}
\renewcommand{\topfraction}{0.92}
\renewcommand{\bottomfraction}{0.80}
\renewcommand{\textfraction}{0.08}
\renewcommand{\floatpagefraction}{0.75}
\renewcommand{\dbltopfraction}{0.92}
\renewcommand{\dblfloatpagefraction}{0.75}
\setcounter{topnumber}{3}
\setcounter{bottomnumber}{2}
\setcounter{totalnumber}{5}

\makeatletter
\setlength{\@fptop}{0pt}
\setlength{\@dblfptop}{0pt}
\makeatother

\twocolumn[{
\centering
\LARGE
GeoFocus: Blending Efficient Global-to-Local Perception for Multimodal Geometry Problem-Solving 

\large (Supplementary Material)
\par
\vspace{1.5em}
}]

\section{Distribution of Critical Local Task Categories per Problem}
\begin{figure}[!t]
    \centering
    \includegraphics[width=1\linewidth]{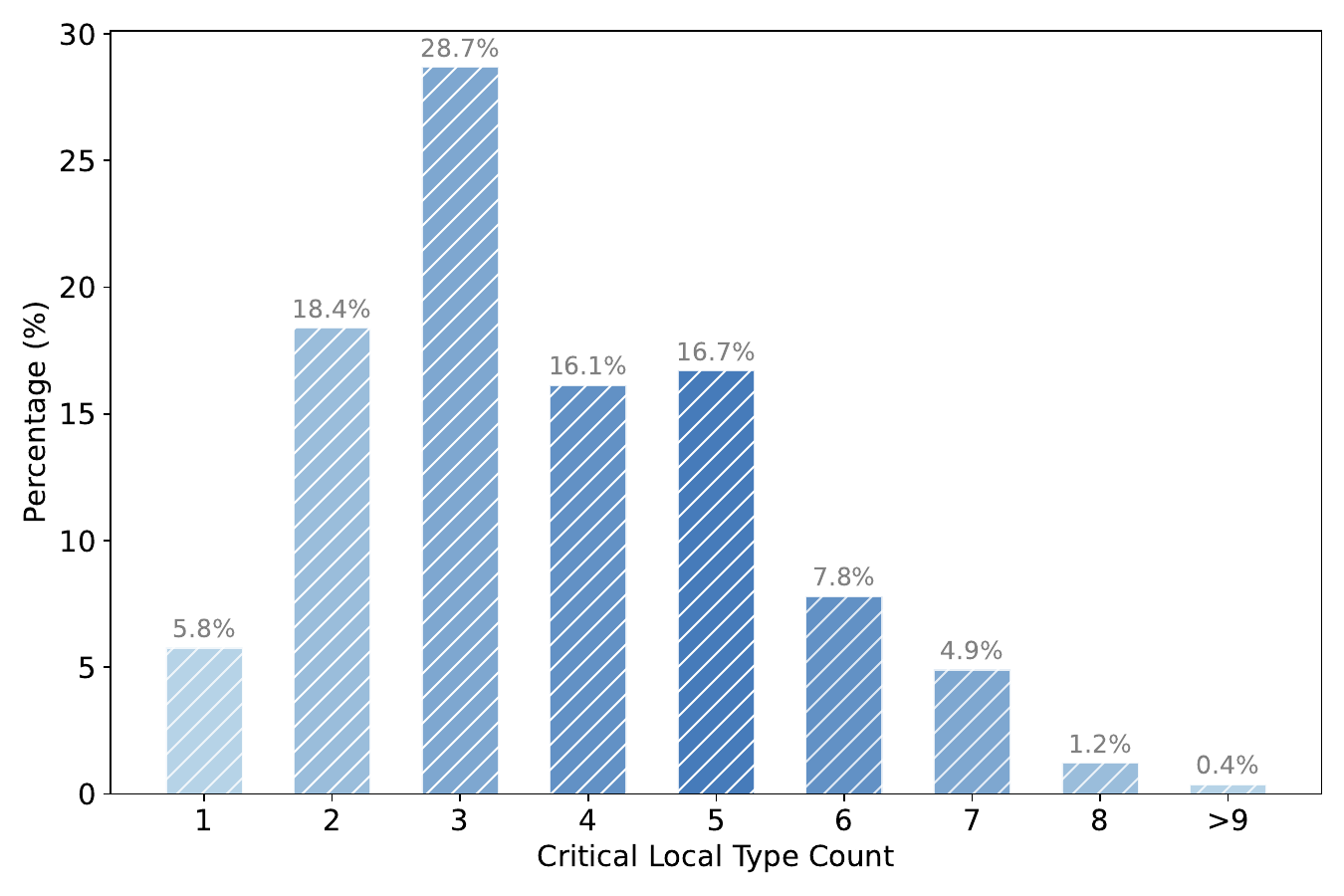}
    \caption{Distribution of critical local task-type counts across 5,000 GPS Q\&A pairs.}
    \label{fig:template_number}
\end{figure}

Fig.~\ref{fig:template_number} illustrates the distribution of the number of critical local task categories involved in 5,000 GPS Q\&A pairs. Pairs involving three types account for the largest proportion (28.7\%), followed by those involving two (18.4\%), five (16.7\%), and four (16.1\%). Overall, 79.9\% of the pairs involve two to five types, indicating that most sampled problems involve multiple annotated local task categories.

\section{Critical Local Structure Templates}
Tab.~\ref{tab:geo_task_templates} provides a summary of the 13 categories of critical local structure templates covered by the proposed Critical Local Perceptor. The table is organized in three columns: Task Type, Question Template, and Answer (including positive and negative forms). The first five types (Angle Value, Angle Compare, Length Value, Length Compare, Area Compare) belong to Basic Measurement and focus on acquiring and comparing quantitative properties; the remaining eight (Shape Check, Bisector Check, Perp. Check, Parallel Check, Midpoint Check, On Circle, Perp. Foot Check, Collinearity) fall under Relational Reasoning and emphasize relationships among points, lines, and shapes. Each template provides both a correct answer and an incorrect answer (separated by a slash), enabling preference-based training. Rejected answers are constructed to be inconsistent with the ground-truth geometric properties.

\begin{table*}[!t]
\centering
\small
\renewcommand\arraystretch{0.94}
\caption{Templates in the Critical Local Perceptor. Chosen and rejected responses are assigned according to the ground-truth geometric properties.}
\setlength{\tabcolsep}{4pt}{
\setlength{\columnwidth}{\linewidth}\resizebox{\columnwidth}{!}{\begin{tabular}{@{}L{0.17\textwidth}L{0.37\textwidth}L{0.42\textwidth}@{}}
\toprule
\textbf{Task Type} & \textbf{Question Template} & \textbf{Answer Variants} \\
\midrule

Angle Value & What is the measure of angle \{A\}? & Angle \{A\} measures \{val\} / \{wrong\_val\} degrees. \\
\addlinespace[0pt]

Angle Compare & Which angle is greater between angle \{A1\} and \{A2\}? & Angle \{A1\} / \{A2\} is greater. Or: The two angles are identical. \\
& Are angles \{A1\} and \{A2\} equal? & Angle \{A1\} is equal to / not equal to angle \{A2\}. \\
\addlinespace[0pt]

Length Value & What is the length of segment \{S\}? & The length of \{S\} is \{val\} / \{wrong\_val\}. \\
& What is the side length of the shape? & The side length is \{val\} / \{wrong\_val\}. \\
& What is the radius of the circle? & The radius is \{val\} / \{wrong\_val\}. \\
\addlinespace[0pt]

Length Compare & Which segment is longer: \{S1\} or \{S2\}? & \{S1\} / \{S2\} is longer. Or: The two segments are equal in length. \\
& Are segments \{S1\} and \{S2\} equal in length? & They are equal / not equal. \\
\addlinespace[0pt]

Area Compare & Which triangle has a greater area: triangle \{T1\} or \{T2\}? & Triangle \{T1\} / \{T2\} has a greater area. \\
& Do triangle \{T1\} and \{T2\} have equal areas? & They have equal / unequal areas. \\
\addlinespace[0pt]

Shape Check & Is the shape formed by \{V\} a \{shape\}? & The shape is / is not a \{shape\}. \\
& What type is quadrilateral \{Q\} (adjacent to \{ref\})? & It is a \{correct\_shape\} / \{wrong\_shape\}. \\
& What type is triangle \{T\}? & It is a \{type\} / \{wrong\_type\}. \\
\addlinespace[0pt]

Bisector Check & Which line is the bisector of angle \{A\}? & Line \{L1\} / \{L2\} is the bisector. \\
               &  Is line \{L\} the bisector of angle \{A\}? & Line \{L\} is / is not the bisector. \\
\addlinespace[0pt]

Perp. Check & Which line is perpendicular to line \{L1\}? & Line \{L3\} / \{L2\} is perpendicular to line \{L1\}. \\
            &  Is line \{L2\} perpendicular to line \{L1\}? & Line \{L2\} is / is not perpendicular to line \{L1\}. \\
\addlinespace[0pt]

Parallel Check & Is line \{L2\} parallel to  \{L1\}? & Line \{L2\} is / is not parallel to line \{L1\}. \\
& Which line is parallel to line \{L1\}? & Line \{L2\} / \{L3\} is parallel to line \{L1\}. \\
\addlinespace[0pt]

Midpoint Check & Is point \{P\} the midpoint of segment \{S\}? & Point \{P\} is / is not the midpoint of segment \{S\}. \\
\addlinespace[0pt]

On Circle & Is point \{P\} on the circle? & Point \{P\} lies on / is not on the circle. \\
\addlinespace[0pt]

Perp. Foot Check & Is point \{P\} the foot of the perpendicular to line \{L\}? & Point \{P\} is / is not the foot. \\
\addlinespace[0pt]

Collinearity & Are points \{P1\}, \{P2\}, \{P3\} collinear? & Points are / are not collinear. \\
\bottomrule
\end{tabular}
}}
\label{tab:geo_task_templates}
\end{table*}

\section{Critical Local Perception Coverage}

We compare the task-category coverage of the Critical Local Perceptor with prior geometry perception Q\&A generation methods, as shown in Tab.~\ref{tab:task-coverage}. For a consistent comparison, we organize 17 local perception task types into three groups: (a) numeric measurements (Angle Value, Length Value, Quantity Check); (b) relative comparisons (Angle Compare, Length Compare, Area Compare, Slope Compare, Position Compare); and (c) structural and constraint relations (Shape Check, Collinearity, Perp. Foot Check, On Circle, Intersect Check, Midpoint Check, Parallel Check, Bisector Check, Perp. Check).
The Critical Local Perceptor covers 13 of the 17 task types (76\%), exceeding CogAlign~\cite{huang2025vision} (47\%), GeoPeP~\cite{sun2025mathglance} (18\%), and MAVIS~\cite{zhang2024mavis} (12\%). Compared with CogAlign, which primarily covers angle and length comparisons and selected relational checks, our method additionally covers structural relations such as collinearity, parallelism, and perpendicularity. This broader task-category coverage provides more diverse supervision of reasoning-relevant geometric relations. Notably, coverage here refers to the proportion of task types in the unified taxonomy, rather than the proportion of GPS problems covered.

\begin{table}[!t]
\centering
\footnotesize
\caption{Comparison of Critical Local Structure Type Coverage Across Different Methods. `CLP' denotes Critical Local Perceptor.}
\renewcommand{\arraystretch}{0.93}
\setlength{\tabcolsep}{3pt}{
\resizebox{\columnwidth}{!}{\begin{tabular}{@{}l c c c c@{}}
\toprule
\textbf{Task Type} & \textbf{CLP} & \textbf{CogAlign} & \textbf{GeoPeP} & \textbf{MAVIS} \\
\midrule
Angle Value     & \ding{51} & \ding{55} & \ding{55} & \ding{55} \\
Length Value    & \ding{51} & \ding{55} & \ding{55} & \ding{51} \\
Quantity Check  & \ding{55} & \ding{51} & \ding{51} & \ding{55} \\
Angle Compare   & \ding{51} & \ding{51} & \ding{55} & \ding{55} \\
Length Compare  & \ding{51} & \ding{51} & \ding{55} & \ding{55} \\
Area Compare    & \ding{51} & \ding{51} & \ding{55} & \ding{55} \\
Slope Compare   & \ding{55} & \ding{51} & \ding{55} & \ding{55} \\
Position Compare    & \ding{55} & \ding{51} & \ding{51} & \ding{55} \\
Shape Check     & \ding{51} & \ding{51} & \ding{51} & \ding{51} \\
Collinearity   & \ding{51} & \ding{55} & \ding{55} & \ding{55} \\
Perp. Foot Check  & \ding{51} & \ding{55} & \ding{55} & \ding{55} \\
On Circle       & \ding{51} & \ding{55} & \ding{55} & \ding{55} \\
Intersect Check & \ding{55} & \ding{51} & \ding{55} & \ding{55} \\
Midpoint Check      & \ding{51} & \ding{55} & \ding{55} & \ding{55} \\
Parallel Check      & \ding{51} & \ding{55} & \ding{55} & \ding{55} \\
Bisector Check      & \ding{51} & \ding{55} & \ding{55} & \ding{55} \\
Perp. Check          & \ding{51} & \ding{55} & \ding{55} & \ding{55} \\
\midrule
\textbf{Task Coverage (\%)} & \textbf{76} & \textbf{47} & \textbf{18} & \textbf{12} \\
\bottomrule
\end{tabular}
}}
\label{tab:task-coverage}
\end{table}

\section{Examples of Critical Local Perception Q\&A Templates and Task Types}

\begin{figure}[!t]
    \centering
    \includegraphics[width=0.95\linewidth]{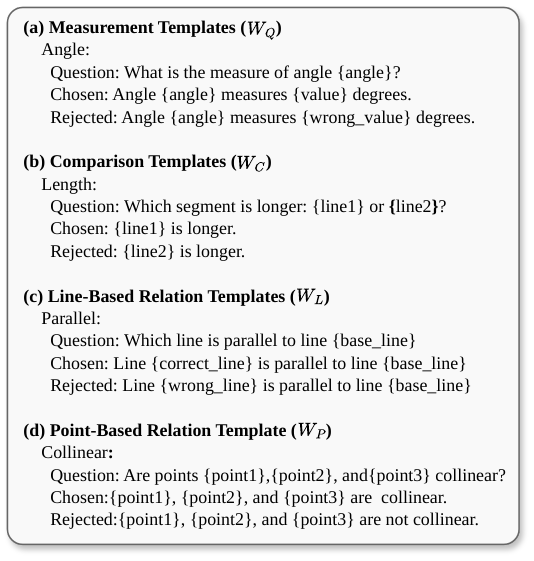}
    \caption{Examples of Critical Local Perception Q\&A Templates}
    \label{fig:template}
\end{figure}

\begin{figure*}[!t]
    \centering
    \includegraphics[width=1\linewidth]{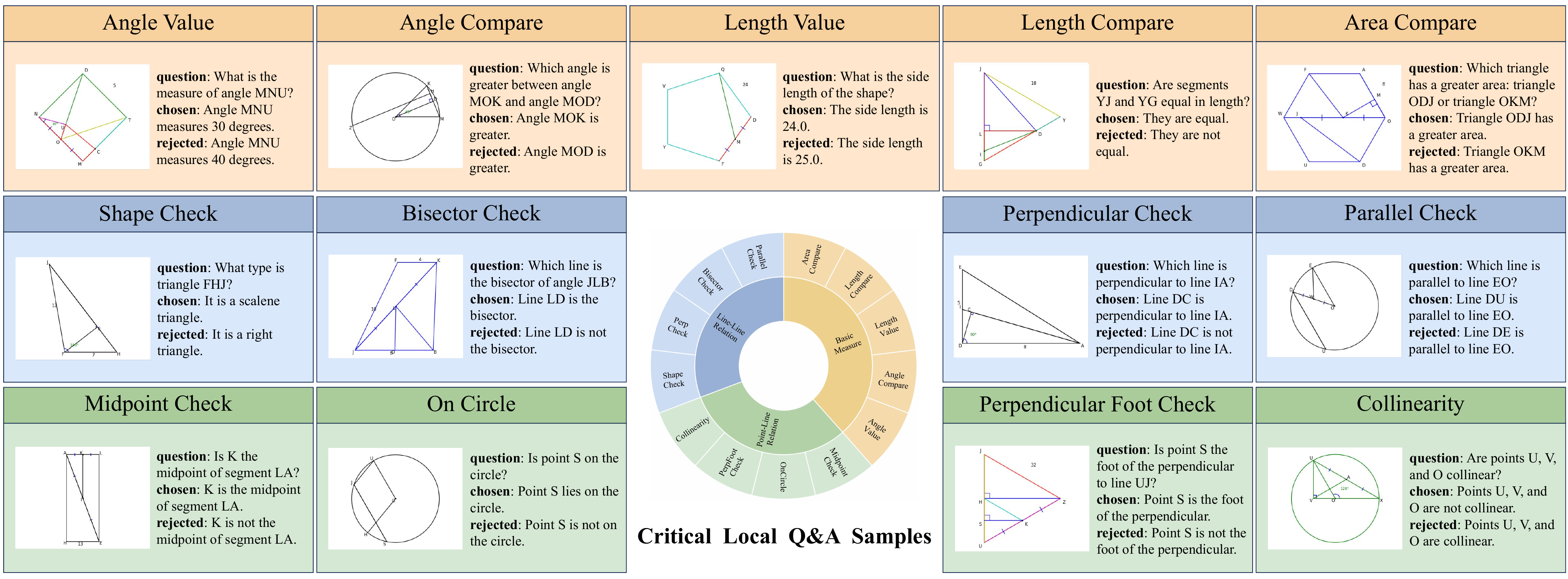}
    \caption{Critical Local Perceptor problem types and typical examples. Each example includes an image, a question, a correct answer aligned with the image, and an incorrect answer that conflicts with it.}
    \label{fig:GeoPercept}
\end{figure*}

Fig.~\ref{fig:template} illustrates four representative template families for generating critical local perception Q\&A pairs, covering angle measurement, length comparison, line–based relations, and point–based relations. Each template instantiates predefined placeholders with diagram-grounded geometric entities and values to construct a question and two contrasting answers: a chosen answer consistent with the diagram and a rejected answer that contradicts it. Fig.~\ref{fig:GeoPercept} further presents representative examples of the 13 perception task types, spanning quantitative measurements and comparisons as well as structural relations among points, lines, and shapes. These examples illustrate how the Critical Local Perceptor transforms fine-grained geometric properties into explicit supervision targets, complementing the global topology reconstruction provided by VertexLang.

\section{Pseudo-code of VertexLang Reconstructor}
Algorithm~\ref{alg:geovision} outlines the construction of geometric diagrams and their corresponding VertexLang representations. It first initializes three data structures: the circle information set ($\mathcal{R}$), the vertex coordinate dictionary ($\mathcal{P}$), and the connectivity dictionary ($\mathcal{E}$). The AddVertex routine generates normalized coordinates, assigns unique vertex annotations, and establishes neighboring relationships according to the specified connectivity constraints ($\mathcal{C}$). The algorithm then samples ($n_1$) geometric substrates from ($\mathcal{S}$), recording circle radii and centers when applicable, and adds their associated vertices and connectivity. Subsequently, ($n_2$) additional vertices are introduced through the same routine. Finally, the geometry renderer ($\mathcal{F}$) takes ($\mathcal{R}, \mathcal{P}, \mathcal{E}$) as input to produce the rendered image (G), while the same structured information is serialized into the VertexLang representation (T). This process ensures that each generated diagram is paired with a compact representation of its geometric coordinates and connectivity.

\begin{algorithm}[!t]
\footnotesize
\SetAlFnt{\footnotesize}
\SetAlCapFnt{\small}
\SetAlCapNameFnt{\small\bfseries}
\caption{VertexLang Reconstructor}
\label{alg:geovision}

\KwIn{Geometry substrate set $\mathbf{S}$, number of substrates $n_1$, number of additional vertices $n_2$, connectivity constraints $\mathcal{C}$}
\KwOut{Rendered geometry image $G$, VertexLang text $T$}

\textbf{Initialization:} \\
radius $\mathbf{R} \gets \emptyset$ \\
vertex dictionary $\mathbf{P} \gets \emptyset$ \\
connectivity dictionary $\mathbf{E} \gets \emptyset$ \\

\textbf{Subroutine: Add Vertex and Connectivity} \\
\SetKwFunction{FAddVertex}{AddVertex}
\SetKwProg{Fn}{Function}{:}{}
\Fn{\FAddVertex{$\mathbf{P}, \mathbf{E}, \mathcal{C}$}}{
    $(x, y) \gets$ generate normalized coordinates in $[0,1]^2$ \\
    $v \gets$ assign unique vertex annotation \\
    $V \gets \mathrm{keys}(\mathbf{P})$ \\
    $\mathbf{P} \gets \mathbf{P} \cup \{v:(x, y)\}$ \\
    $\mathcal{N}(v) \gets \{u \in V \mid u \sim v\}$ based on $\mathcal{C}$ \\
    $\mathbf{E}[v] \gets \mathcal{N}(v)$ \\
    \ForEach{$u \in \mathcal{N}(v)$}{
        $\mathbf{E}[u] \gets \mathbf{E}[u] \cup \{v\}$ \\
    }
}

\textbf{Step 1: Substrate Selection} \\
\For{$i \gets 1$ \KwTo $n_1$}{
    $s_i \gets$ sample from $\mathbf{S}$ \\
    \If{$s_i$ is a circle}{
        $(o_i, r_i) \gets$ extract circle center and radius from $s_i$ \\
        $\mathbf{R} \gets \mathbf{R} \cup \{o_i:r_i\}$ \\
        \If{$o_i \notin \mathrm{keys}(\mathbf{P})$}{
            $(x_{o_i},y_{o_i}) \gets$ normalized center coordinates \\
            $\mathbf{P} \gets \mathbf{P} \cup \{o_i:(x_{o_i},y_{o_i})\}$ \\
            $\mathbf{E}[o_i] \gets \emptyset$ \\
        }
    }
    \FAddVertex{$\mathbf{P}, \mathbf{E}, \mathcal{C}$}
}

\textbf{Step 2: Additional Vertex and Connectivity Generation} \\
\For{$i \gets 1$ \KwTo $n_2$}{
    \FAddVertex{$\mathbf{P}, \mathbf{E}, \mathcal{C}$}
}

\textbf{Step 3: Geometry Rendering} \\
$G \gets \mathcal{F}(\mathbf{R}, \mathbf{P}, \mathbf{E})$ \\
$T \gets (\mathbf{R}, \mathbf{P}, \mathbf{E})$

\Return{$G, T$}
\end{algorithm}

\section{Benchmarks and Evaluation Protocols}
\textbf{Benchmarks.}
Experiments were conducted on three widely used multimodal GPS benchmarks: Geometry3K (Geo3K)~\cite{lu2021inter}, GeoQA~\cite{chen2021geoqa}, and FormalGeo7K~\cite{zhang2023formalgeo}. Below is a brief overview of each test dataset:
1) GeoQA consists of 754 geometry questions sourced from Chinese secondary school exams. The images in this dataset are directly obtained by scanning exam papers, reflecting the visual characteristics of real-world GPS tasks. This dataset evaluates geometric reasoning on scanned exam diagrams.
2) Geo3K contains 601 geometry questions derived from example problems in secondary school textbooks, with manually annotated questions and answer options. The textual content of Geo3K is relatively concise, and most critical information for problem-solving is embedded in geometric images. Consequently, this dataset poses higher demands on a model's visual perception abilities. 
3) FormalGeo7K comprises 1,050 geometry problems collected from online resources and curated manually to ensure quality. It features a broad range of problem types to provide a comprehensive evaluation setting for GPS models.

We evaluated models' geometric perception using the CogAlign-data~\cite{huang2025vision} dataset, which comprises synthetic images paired with Q\&A pairs across four core tasks probing geometric perception.
Additionally, out-of-domain generalization was evaluated on four prominent multimodal mathematics datasets (We-Math~\cite{qiao2025we}, MathVerse~\cite{zhang2024mathverse}, MathVision~\cite{wang2024measuring}, and MathVista~\cite{lu2023mathvista}), as well as the hallucination dataset HallusionBench~\cite{guan2024hallusionbench} and the chart question answering benchmark ChartQA~\cite{masry2022chartqa}.

\textbf{Evaluation and Metrics.}
To evaluate the model's reasoning ability on the GPS dataset, we employ the EasyR1~\cite{zheng2025easyr1} format script to extract open-ended answers and compute accuracy metrics. Total denotes the sum of Top-1 accuracies across Geo3K, GeoQA, and FormalGeo7K. For the out-of-domain datasets, we utilize the evaluation scripts provided by NoisyRollout~\cite{liu2026noisyrollout} to assess the accuracy.
For in-domain topology reconstruction, we directly evaluate the accuracy of predicted vertex coordinates and connectivity against ground-truth VertexLang annotations. For out-of-domain evaluation, we render reconstructed outputs into images and convert them into structured representations using FGeoParser~\cite{zhu2024fgeo}, enabling cross-format comparisons through BLEU-4 and Construction CdlAcc~\cite{zhu2024fgeo}.

\textbf{Model Backbones.} Unless otherwise specified, the training-strategy and module-order ablations, topology reconstruction evaluation, and qualitative reasoning examples use Qwen2.5-VL-3B and Qwen2.5-VL-7B as the base models. We additionally evaluate GeoFocus on Gemma3-4B and Qwen3-VL-2B/4B to investigate its effectiveness across different model architectures and scales.

\section{DynamicGT-RL}
During training using VertexLang-based reconstructed samples generated by the VertexLang Reconstructor (Tab.~\ref{topo_training_algorithm}), we observed that supervised fine-tuning effectively injects task-specific knowledge but weakens the model's generalization ability, limiting the application of perceptual skills from reconstruction tasks to reasoning tasks, and resulting in performance degradation when transferring topology perception to reasoning tasks. Reinforcement learning (RL) algorithms, while preserving reasoning generalization ability, provide limited improvement due to the model's weak initial performance in image reconstruction, which makes RL struggle to identify optimization directions~\cite{yue2026does}. To accelerate convergence and improve training stability in the early stages, we propose DynamicGT-RL, which probabilistically incorporates ground-truth samples into group-relative advantage estimation.

Specifically, for each question $q$, DynamicGT-RL samples a group of outputs $\{o_1,o_2,...,o_U\}$ from the old
rollout $\pi_{\theta_{old}}$ and then optimizes the policy model $\pi_{\theta}$ by maximizing the following objective:

\begin{equation}
\begin{split}
    \mathcal{J}(\theta) &= \mathbb{E}_{q \sim P(Q), \{o_i\}_{i=1}^U \sim \pi_{\theta_{old}}(O|q)} \\
    &\quad \Bigg[\frac{1}{|\mathcal{P}|}\sum_{o_i\in\mathcal{P}}
        \min \Bigg( 
            \frac{\pi_\theta(o_i |q)}{\pi_{\theta_{old}}(o_i |q)} \hat{A_i}(R) , \\
            &\quad\quad \text{clip} \Big( 
                \frac{\pi_\theta(o_i |q)}{\pi_{\theta_{old}}(o_i |q)}, 
                1 - \epsilon, 
                1 + \epsilon 
            \Big) \hat{A_i}(R) 
        \Bigg) \\
        &\quad - \beta \mathbb{D}_{KL}\Big(\pi_{\theta} || \pi_{ref}\Big)
    \Bigg],
\end{split}
\label{eq:stage1}
\end{equation}
\begin{equation}
\begin{aligned}
    &\mathbb{D}_{KL}\left(\pi_{\theta} || \pi_{ref}\right) \\
    &\quad= \frac{\pi_{ref}(o_i|q)}{\pi_{\theta}(o_i|q)}
      - \log\frac{\pi_{ref}(o_i|q)}{\pi_{\theta}(o_i|q)} - 1,
\end{aligned}
\end{equation}
where $\epsilon$ and $\beta$ are hyper-parameters. $\mathcal{P}=\mathcal{P}_{\mathrm{model}}\cup
\mathcal{P}_{\mathrm{gt}}$ when $x<\tau$, and
$\mathcal{P}=\mathcal{P}_{\mathrm{model}}$ otherwise. $\pi_{\theta_{old}}(O|q)$ does not participate in gradient updates. $\hat{A_i}(R)$ denotes the advantage and $R$ represents the reward function. The incorporation of GT influences $\hat{A_i}(R)$ calculation in the following ways:

\begin{equation}
\begin{aligned}
&\hat{A}_i(R(P))= \\
&\begin{cases}
\begin{aligned}
&\displaystyle\frac{R\left(p_i\right)-\text { mean }\left(R\left(P_{\text {model }}\right)\right)}{\operatorname{std}\left(R\left(P_{\text {model }}\right)\right)},\\[-1pt]
&\hspace{1em}\text { if } x \geq \tau ,
\end{aligned}\\[5pt]
\begin{aligned}
&\displaystyle\frac{R\left(p_i\right)-\operatorname{mean}\left(R\left(P_{\text {model }} \cup P_{\text {gt }}\right)\right)}{\operatorname{std}\left(R\left(P_{\text {model }} \cup P_{\text {gt }}\right)\right)},\\[-1pt]
&\hspace{1em}\text { if } x<\tau ,
\end{aligned}
\end{cases}
\end{aligned}
\label{app:eq:ADV}
\end{equation}
where $p_i$ is a rollout, $P$ is a set of rollouts, $x$ is a random number between 0 and 1, and $\tau$ controls the probability of including GT samples in advantage estimation.
When the model fails to generate correct reconstructions, the GT response receives a higher advantage and provides a positive optimization signal. As the model improves and begins generating correct solutions, the advantage of the GT rollout gradually decreases, allowing the model to rely more on its own rollouts without excessive dependence on the GT.

For the VertexLang-based image reconstruction task, we carefully design a Recon Reward $R_{\text {recon }}$, which consists of two verifiable reward functions: vertex coordinate accuracy reward $R_{\text {coord}}$ and connectivity relationship accuracy reward $R_{conn}$. Vertex coordinate accuracy reward is assessed by calculating the Euclidean distance between the model-predicted coordinates and the corresponding GT coordinates. Based on the distance $d(v)$ and predefined thresholds $\tau_1$ and $\tau_2$, the score for each vertex is defined as follows:
\begin{equation}
\operatorname{score}(v)= \begin{cases}1.0, & \text { if } d(v) \leq \tau_1 \\ 0.5, & \text { if } \tau_1<d(v) \leq \tau_2 \\ 0.0, & \text { if } d(v)>\tau_2\end{cases}
\end{equation}
The final vertex coordinate accuracy score is the average of all vertex scores:
\begin{equation}
R_{\text {coord}}=\frac{1}{|\mathcal{V}|} \sum_{v \in \mathcal{V}} \operatorname{score}(v),
\end{equation}
where $|\mathcal{V}|$ denotes the total number of vertices.
Connectivity relationship accuracy reward is measured using the F1 score $R_{conn}$, which compares the model-predicted edge set $E_{model}$ with the GT edge set $E_{gt}$. The final reward score is computed by the vertex coordinate accuracy score and the connectivity relationship accuracy score:
\begin{equation}
R_{\text {recon }}=0.5*R_{\text {coord }} + 0.5*R_{\text {conn }}.
\end{equation}
where $R_{recon}$ is substituted into Eq.~\ref{app:eq:ADV} to perform the corresponding calculation.

As shown in Tab.~\ref{topo_training_algorithm}, supervised learning reduces downstream GPS performance relative to the original backbone, while standard RL yields only modest improvements. These observations motivate DynamicGT-RL, which incorporates GT guidance into advantage estimation to facilitate reconstruction training.

\begin{table}[!t]
\centering
\footnotesize
\caption{Image Reconstruction Training Method Ablation Experiment. `SFT' represents Supervised Learning. `RL' represents Reinforcement Learning. `GT-RL' represents DynamicGT-RL}
\setlength{\tabcolsep}{3pt}{
\resizebox{\columnwidth}{!}{\begin{tabular}{@{}c l c c c c@{}}
\toprule 
& Training     & \multicolumn{1}{c}{Geo3K} & \multicolumn{1}{c}{GeoQA}    & \multicolumn{1}{c}{FormalGeo7K}  & \multicolumn{1}{c}{Total}\\ \midrule
\multirow{4}{*}{3B} & Baseline  &25.3  &32.1  &16.8  &74.2 \\
      & + SFT                &  25.4                &  31.6    & 14.9  &  71.9           \\
           & + RL                &  25.3                &  32.4     &  16.7   &  74.4            \\
         & + GT-RL                 &  25.4               &  33.8       &  17.3   &  76.5           \\ \midrule

\multirow{4}{*}{7B}   
& Baseline  &39.4  &44.4  &35.6  &119.4 \\
       & + SFT                               & 38.9   & 43.5   & 34.4   & 116.8         \\
            & + RL                &  39.6                &  46.3      &  35.0   & 120.9            \\
       & + GT-RL                 &  39.8              &  46.6       &  36.0   &  122.4          \\ \bottomrule

\end{tabular}
}}
\label{topo_training_algorithm}
\end{table}

\section{Ablation of Local Perception Q\&A Training Strategies}

We compare supervised fine-tuning (SFT) and DPO for critical local perception Q\&A training (Tab.~\ref{align_training_algorithm}). DPO consistently outperforms SFT across all three GPS benchmarks on both 3B and 7B backbones, suggesting that preference-based supervision is more effective for learning reasoning-relevant local geometric relations.

\begin{table}[!t]
\centering
\caption{Perception Q\&A Training Strategy Ablation Experiment. `SFT' represents Supervised Learning.}
\footnotesize
\setlength{\tabcolsep}{3pt}{
\resizebox{\columnwidth}{!}{\begin{tabular}{@{}c l c c c c@{}}
 \toprule
& Training      & \multicolumn{1}{c}{Geo3K} & \multicolumn{1}{c}{GeoQA}    & \multicolumn{1}{c}{FormalGeo7K}  & \multicolumn{1}{c}{Total}\\ \midrule
    \multirow{2}{*}{3B} & SFT                 &  25.5                   &  33.4       &  17.0   & 75.9           \\
     & DPO                 &  25.6                      &  34.2       &  17.3   &  77.1         \\ \midrule
     
\multirow{2}{*}{7B}    
 & SFT                 &  36.6                                  &  44.3       &  33.7   & 114.6             \\
     & DPO                 &  40.1                      &  46.2       &  35.6   &  121.9         \\ \bottomrule
\end{tabular}
}}
\label{align_training_algorithm}
\end{table}

\section{Effect of Module Order on GPS Performance} 
\label{sec:order}

To examine the effect of module ordering, we compare global-to-local and local-to-global training sequences. As shown in Tab.~\ref{order}, applying the VertexLang Topology Perceptor before the Critical Local Perceptor improves the summed accuracy by 4.4 and 1.3 points on the 3B and 7B backbones, respectively. These results support the global-to-local training sequence adopted in GeoFocus.

\begin{table}[!t]
\centering
\caption{Exploring the sequential dependency between the proposed modules. `1' denotes the VertexLang Topology Perceptor. `2' denotes the Critical Local Perceptor.}
\footnotesize
\setlength{\tabcolsep}{3pt}{
\resizebox{\columnwidth}{!}{\begin{tabular}{@{}l c c c c@{}}
\toprule
\multicolumn{1}{c}{Order}  & \multicolumn{1}{c}{Geo3K}  & \multicolumn{1}{c}{GeoQA}       & \multicolumn{1}{c}{FormalGeo7K}  & \multicolumn{1}{c}{Total}  \\ \midrule
   3B  2$\rightarrow$1        & 26.5  & 32.9    & 15.8   & 75.2 \\
3B  1$\rightarrow$2    & 26.5  & 34.4     & 18.7  & 79.6 (4.4$\uparrow$) \\ 
7B  2$\rightarrow$1  & 39.3  & 47.0    & 35.9  & 122.2\\
  7B 1$\rightarrow$2   & 40.3  & 47.1    & 36.1   & 123.5 (1.3$\uparrow$) \\ \bottomrule
\end{tabular}
}}
\label{order}
\end{table}

\section{Evaluation on an Additional Backbone}
\label{sec:baseline}
We further validated the effectiveness of \mymethod{} on Gemma3-4B~\cite{team2025gemma}. As shown in Tab.~\ref{otherbase}, integrating either module independently or combining both modules consistently led to performance improvements. These results provide additional evidence of the effectiveness of \mymethod{} on Gemma3-4B, suggesting that the proposed modules are not limited to a single model backbone.

\begin{table}[!t]
\centering
\caption{Ablation study of \mymethod{} on Gemma3-4B.
`Topo' and `Local' denote the VertexLang Topology Perceptor and Critical Local Perceptor, respectively. `FGeo7K' denotes the FormalGeo7K.
Both modules are trained on synthetic data.}
\label{otherbase}

\footnotesize
\setlength{\tabcolsep}{3pt}
\renewcommand{\arraystretch}{1.15}

\resizebox{\columnwidth}{!}{\begin{tabular}{@{}l c c c c c c@{}}
\toprule
Model & Topo & Local & Geo3K & GeoQA & FGeo7K & Total \\
\midrule
\multirow{4}{*}{Gemma3-4B}
& \ding{55} & \ding{55} & 30.6 & 42.6 & 37.0 & 110.2 \\
& \ding{51} & \ding{55} & 31.6 & 43.8 & 38.9 & 114.3 {\scriptsize($\uparrow$4.1)} \\
& \ding{55} & \ding{51} & 31.8 & 44.3 & 38.4 & 114.5 {\scriptsize($\uparrow$4.3)} \\
& \ding{51} & \ding{51} & \textbf{33.3} & \textbf{45.6} & \textbf{40.7}
& \textbf{119.6} {\scriptsize($\uparrow$9.4)} \\
\bottomrule
\end{tabular}
}\end{table}

\section{Geometric Topology Reconstruction Performance}
\label{Reconstruction-Performance-Comparison}
To explore the effectiveness of VertexLang in enhancing topology perception and reconstruction generalization, we compare in-domain and out-of-domain topology reconstruction performance. The in-domain test set comprises 500 VertexLang-based reconstruction instances, separate from the training set. As shown in the In-Domain section of Tab.~\ref{indomain_Recon}, VertexLang training improves connection accuracy (Connect. Acc.) and coordinate accuracy (Coord. Acc.) by an average of 20.3 and 17.5 percentage points (pp), respectively.

Since the in-domain test set provides ground-truth annotations only in VertexLang, we conduct cross-format comparisons on out-of-domain data, where both VertexLang and CodeLang outputs are evaluated using the same rendering and parsing pipeline. Specifically, we use the FormalGeo7K test set, which contains geometry diagrams outside the synthetic reconstruction training set. As shown in the Out-of-Domain section of Tab.~\ref{indomain_Recon}, CodeLang yields lower structural accuracy (Cons. CdlAcc) than the original backbones. In contrast, VertexLang improves BLEU-4 by 1.4/9.6 points and Construction CdlAcc by 2.6/7.8 pp over the original Qwen2.5-VL-3B/7B backbones, respectively. These results demonstrate the effectiveness of compact topology reconstruction supervision, with improvements extending to unseen diagrams from FormalGeo7K.

\begin{table}[!t]
\centering
\caption{Comparison of image reconstruction ability on in-domain and out-of-domain data.}
\footnotesize
\setlength{\tabcolsep}{3pt}{
\resizebox{\columnwidth}{!}{\begin{tabular}{@{}c l c c@{}}
\toprule
&Model  & Connect. Acc. & Coord. Acc. \\  \midrule
\multirow{4}{*}{\makecell{In-\\Domain}} &   Qwen2.5-VL-3B        & 41.9  & 8.9 \\
&  + VertexLang    & 62.5  & 30.9 \\
&  Qwen2.5-VL-7B   & 48.9  & 38.7 \\ 
&  + VertexLang  & 68.9  & 51.6 \\ \midrule
&Model  & BLEU-4 & Cons. CdlAcc \\  \midrule
\multirow{6}{*}{\makecell{Out-of-\\Domain}} &   Qwen2.5-VL-3B        & 32.6  & 12.8 \\
&  + CodeLang    & 33.6  & 3.3 \\
&  + VertexLang    & 34.0  & 15.4 \\
&  Qwen2.5-VL-7B   & 38.1  & 30.1 \\  
&  + CodeLang    & 33.0  & 4.2 \\
&  + VertexLang  & 47.7  & 37.9 \\ \bottomrule
\end{tabular}
}}
\label{indomain_Recon}
\end{table}

\section{Effect of VertexLang Reconstruction Targets on Downstream GPS Performance}

We investigate whether generating complete Python rendering code provides additional benefits over compact reconstruction targets. Specifically, both VertexLang and RenderCode describe geometric diagrams using vertex coordinates and connectivity, while RenderCode additionally includes procedural code required for image rendering. We evaluate both variants using Qwen3-VL-2B and Qwen3-VL-4B on Geo3K, GeoQA, and FormalGeo7K. Both variants use the same model initialization, training samples, and optimization settings, differing only in their reconstruction targets.

As shown in Tab.~\ref{reasoning_recon}, VertexLang consistently outperforms RenderCode across all three benchmarks and both model sizes, improving the summed accuracy by 2.6 and 2.4 pp for the 2B and 4B models, respectively. These results suggest that compact reconstruction targets can provide effective topology supervision without requiring models to learn additional rendering procedures. By avoiding unnecessary procedural details, VertexLang offers a simpler supervision target while achieving better downstream GPS performance.

\begin{table}[!t]
\centering
\footnotesize
\renewcommand{\arraystretch}{1.15}
\setlength{\tabcolsep}{3pt}
\caption{Ablation of VertexLang reconstruction targets on downstream GPS performance using Qwen3-VL-2B and Qwen3-VL-4B.}
\resizebox{\columnwidth}{!}{\begin{tabular}{@{}c l c c c c@{}}
\toprule
Model & Target & Geo3K & GeoQA & FormalGeo7K & Total \\
\midrule
2B & VertexLang
& \textbf{31.5} & \textbf{42.3} & \textbf{34.3} & \textbf{108.1} \\
   & RenderCode
& 30.1 & 41.8 & 33.6 & 105.5 \\
\midrule
4B & VertexLang
& \textbf{50.8} & \textbf{64.3} & \textbf{56.3} & \textbf{171.4} \\
   & RenderCode
& 50.6 & 62.7 & 55.7 & 169.0 \\
\bottomrule
\end{tabular}
}\label{reasoning_recon}
\end{table}

\section{Qualitative Reasoning Examples}
\begin{figure*}[!t]
    \centering
    \includegraphics[width=1\linewidth]{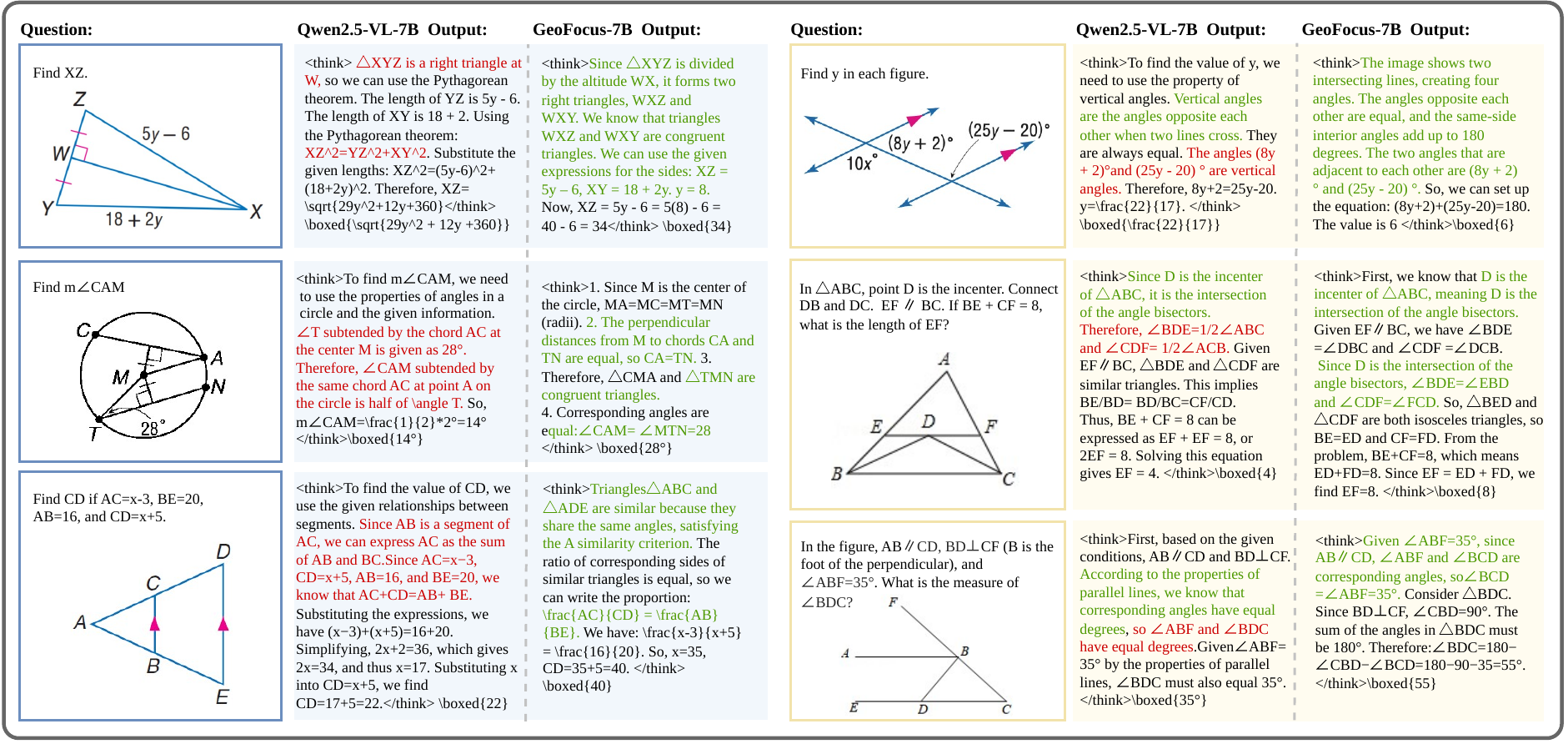}
    \caption{Qualitative example of reasoning. \mymethod-7B produces more reliable multi-step reasoning responses by jointly enhancing global topology understanding and critical local perception. Red and green indicate incorrect and correct final answers, respectively.
    }
    \label{fig:Qualitative_results}
\end{figure*}

To better understand the performance improvements introduced by \mymethod, we conducted a qualitative comparison between the baseline model and our \mymethod-7B model. Fig.~\ref{fig:Qualitative_results} illustrates reasoning examples across different types of geometric images. The baseline model exhibits a primary limitation in GPS: it lacks an integrated perception of global geometric topology and critical local structures. This limitation leads to misinterpretations of triangle types, circle angle classifications, or collinear relationships (as shown in the left example of Fig.~\ref{fig:Qualitative_results}); and also causes failures in grounding geometric theories in concrete visual contexts (e.g., correctly identifying vertical angles, as in the first example on the right side of Fig.~\ref{fig:Qualitative_results}), even when the underlying definitions are recalled. In contrast, the responses generated by \mymethod-7B demonstrate an accurate understanding of global topology and critical local structures, resulting in more accurate and reliable GPS performance.

\section{Inference Efficiency}

We compare the end-to-end inference efficiency of GeoFocus-8B and GeoWeave-8B on Geo3K, GeoQA, and FormalGeo7K under identical hardware conditions, with a batch size of 1. Under their respective default inference settings, GeoFocus-8B completes all three benchmarks in 3 h 50 min, compared with 44 h 30 min for GeoWeave-8B, achieving an 11.6$\times$ speedup without external tools or test-time scaling.

\section{Limitations}
In point-circle positional cases, limited coordinate precision may cause points intended to lie on a circle to deviate slightly from its circumference, introducing geometric inconsistencies during rendering. VertexLang prioritizes vertex coordinates when such inconsistencies arise. To mitigate their downstream impact, the Critical Local Perceptor includes an On Circle task that explicitly supervises point-circle relations, although it does not directly eliminate reconstruction errors.

\section{Future Work}
GeoFocus combines theory-based local templates with the concise formal language VertexLang to improve the accuracy of planar geometric reasoning. Its coordinate-and-connectivity representation also suggests a potential extension to three-dimensional geometry. Future work will focus on: (1) extending VertexLang to effectively represent three-dimensional geometric structures, including vertices, edges, and faces, thereby enabling modeling of spatial relationships; and (2) enhancing the Critical Local Perceptor to process three-dimensional visual perception for accurate reasoning in three-dimensional environments. 
Additionally, integrating spatial reasoning and volumetric analysis into the framework could unlock new possibilities for solving advanced geometric challenges, such as the analysis of polyhedra, curved surfaces, and transformations in three-dimensional space.

\end{document}